%% file: main.tex
\documentclass{svjour3}

\usepackage[utf8]{inputenc}
\usepackage[T1]{fontenc}
\usepackage{lmodern}
\usepackage{pgfplots}
\usepackage{tikz}
\usepackage{microtype}
\usepackage{subcaption}
\usepackage{fancyvrb}
\usepackage{graphicx}
\usepackage{algorithm}
\usepackage{algorithmic}
\usepackage{epsfig}
\usepackage{amsmath}
\usepackage{wrapfig}

\usetikzlibrary{plotmarks}

\makeatletter
\newcommand\footnoteref[1]{\protected@xdef\@thefnmark{\ref{#1}}\@footnotemark}
\makeatother

\pgfplotsset{compat=1.10}

\journalname{Machine learning}

\begin{document}

\title{Beneficial and Harmful Explanatory Machine Learning}

\author{Lun Ai \and Stephen H. Muggleton \and C\'{e}line Hocquette \and Mark Gromowski \and Ute Schmid }


\institute{Lun Ai \at
           Department of Computing, Imperial College London, London, UK\\
           \email{lun.ai15@imperial.ac.uk}
           \and
           Stephen H. Muggleton \at
           Department of Computing, Imperial College London, London, UK\\
           \email{s.muggleton@imperial.ac.uk}
           \and
           C\'{e}line Hocquette \at
           Department of Computing, Imperial College London, London, UK\\
           \email{celine.hocquette16@imperial.ac.uk}
           \and
           Mark Gromowski\at
           Cognitive Systems Group, University of Bamberg, Bamberg, Germany\\
           \email{mark.gromowski@uni-bamberg.de}
           \and
           Ute Schmid\at
           Cognitive Systems Group, University of Bamberg, Bamberg, Germany\\
           \email{ute.schmid@uni-bamberg.de}
}

\date{Received: date / Accepted: date}

\maketitle
\input{00-abstract}
\input{01-intro}

\input{02-related}

\input{03-theoryframework}

\input{04-expframework}

\input{05-experiments}

\input{06-discussion}
\input{07-conclusions}
\input{08-acknowledgements}

\bibliographystyle{abbrv}
\bibliography{cognitive,framework,relatedworkbib,relatedworkaibib,materials,concl}
\end{document}

%% file: 00-abstract.tex
\begin{abstract}
Given the recent successes of Deep Learning in AI there has been increased interest in the role and need for explanations in machine learned theories. A distinct notion in this context is that of Michie's definition of Ultra-Strong Machine Learning (USML). USML is demonstrated by a measurable increase in human performance of a task following provision to the human of a symbolic machine learned theory for task performance. A recent paper demonstrates the beneficial effect of a machine learned logic theory for a classification task, yet no existing work to our knowledge has examined the potential harmfulness of machine's involvement for human comprehension during learning. This paper investigates the explanatory effects of a machine learned theory in the context of simple two person games and proposes a framework for identifying the harmfulness of machine explanations based on the Cognitive Science literature. The approach involves a cognitive window consisting of two quantifiable bounds and it is supported by empirical evidence collected from human trials. Our quantitative and qualitative results indicate that human learning aided by a symbolic machine learned theory which satisfies a {\em cognitive window} has achieved significantly higher performance than human self learning. Results also demonstrate that human learning aided by a symbolic machine learned theory that fails to satisfy this window leads to significantly worse performance than unaided human learning.

\end{abstract}

%% file: 01-intro.tex
\newpage
\section{Introduction}
\label{intro}

In a recent paper \cite{US2018} the authors provided an operational definition for comprehensibility of logic programs and used this, in experiments with humans, to provide the first demonstration of Michie's {\em Ultra-Strong Machine Learning} (USML). The authors demonstrated USML via empirical evidence that humans improve out-of-sample performance in concept learning from a training set $E$ when presented with a first-order logic theory which has been machine learned from $E$. The improvement of human performance indicates a beneficial effect of comprehensible machine learned models on human skill acquisition. The present paper investigates the explanatory effects of machine's involvement in human skill acquisition of simple games. In particular, we have focused on a two-player game as the material for experimentation which was designed to be isomorphic to Noughts and Crosses but features a different spatial arrangement of the game. Our results indicate that when a machine learned theory is used to teach strategies to humans in a noise-free setting, in some cases the human's out-of-sample performance is reduced. This degradation of human performance is recognised to indicate the existence of harmful explanations. 

\begin{figure}[H]
	\centering
	\includegraphics[width=\textwidth]{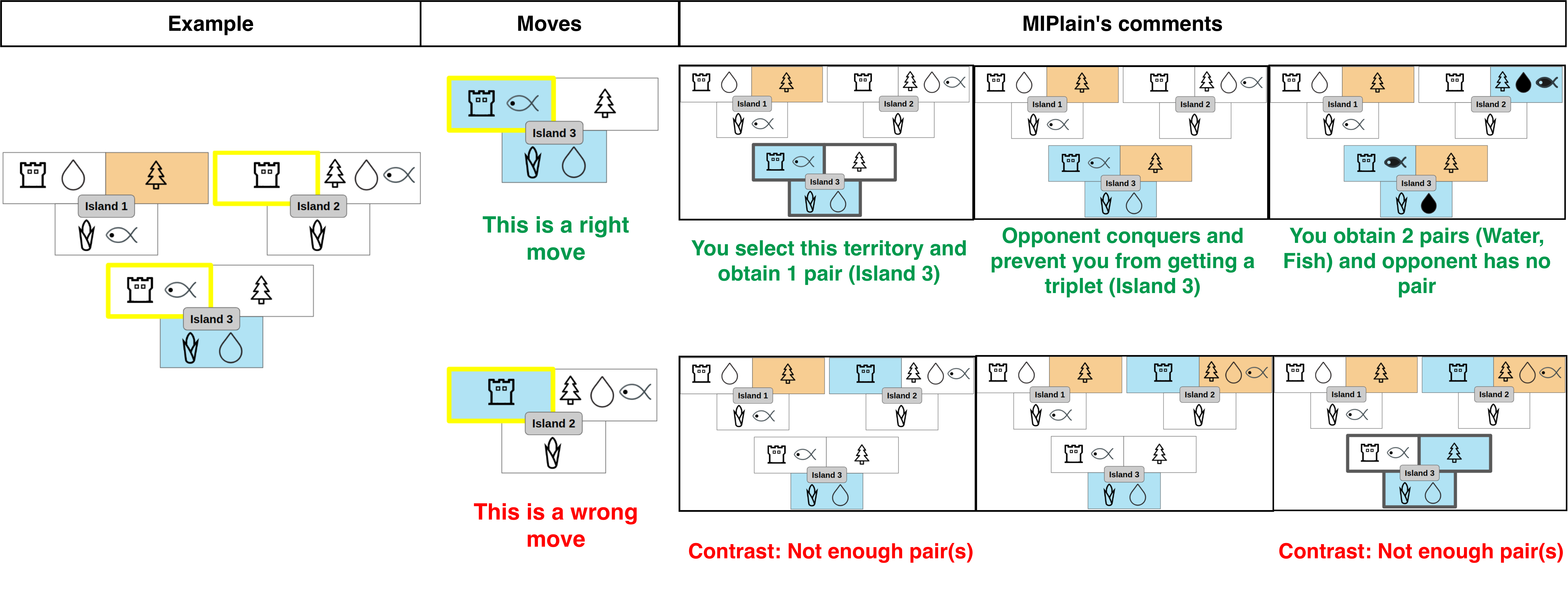}
	\caption{Interface featuring an example of the Island Game that is isomorphic to Noughts and Crosses. Players occupy cells in turns which have resources marked as symbols and a player wins if he or she controls three cells on the same island or three pieces of the same resource. Human participants, who play Blue, are confronted with a game position and have to choose between two alternative moves that are highlighted in yellow. When Blue owns two cells on the same island or two pieces of the same resource, related cells or resources are highlighted in bold. More details of the material design are given in Section \ref{materials}.} Textual and visual explanations\protect\footnotemark are shown to treated participants along with a training example for winning a two player game isomorphic to Noughts and Crosses. Textual explanations were generated from the rules learned by our \textbf{M}eta-\textbf{I}nterpretive ex\textbf{Plain}able game learner $MIPlain$. 
	\label{fig:explanation}
	\vspace{-15pt}
\end{figure}

\footnotetext{Fonts and figures may look larger compared to the actual interface for visual clarity of the paper.} In the current paper, which extends our previous work on the phenomenon of USML, both beneficial and harmful effects of a machine learned theory are explored in the context of simple games. Our definition of explanatory effects is based on human out-of-sample performance in the presence of natural language and visual explanation generated from a machine learned theory (Figure \ref{fig:explanation}). The analogy between understanding a logic program via declarative reading and understanding a piece of natural language text allows the explanatory effects of a machine learned theory to be investigated. 

The results of relevant Cognitive Science literature allow the properties of a logic theory which are harmful to human comprehension to be characterised. Our approach is based on developing a framework describing a cognitive window which involves bounds with regard to 1) descriptive complexity of a theory and 2) execution stack requirements for knowledge application. We hypothesise that a machine learned theory provides a harmful explanation to humans when theory complexity is high and execution is cognitively challenging. Our proposed cognitive window model is confirmed by empirical evidence collected from multiple experiments involving human participants of various backgrounds. \\

\vspace{-5pt}
\noindent We summarise our main contributions as follows:
\vspace{-2pt}
\begin{itemize}
    \item We define a measure to evaluate beneficial/harmful
	explanatory effects of machine learned theory on human comprehension. 
	\item We develop a framework to assess a cognitive window of a machine learned theory. The approach encompasses theory complexity and the required execution stack.
	\item Our quantitative and qualitative analyses of the experimental results demonstrate that a machine learned theory has a harmful effect on human comprehension when its search space is too large for human knowledge acquisition and it fails to incorporate executional shortcuts.
\end{itemize}

This paper is arranged as follows. In Section \ref{related}, we discuss existing work relevant to the paper. The theoretical framework with relevant definitions is presented in Section \ref{framework}. We describe our experimental framework and the experimental hypotheses in Section \ref{exp_framework}. Section \ref{experiments} describes several experiments involving human participants on two simple games. We examine the impact of a cognitive window on the explanatory effects of a machine learned theory based on human performance and verbal input. In Section \ref{conclusion}, we conclude our work and comment on our analytical results -- only a short and simple-to-execute theory can have a beneficial effect on human comprehension. We discuss potential extensions to the current framework, curriculum learning and behavioural cloning, for enhancing explanatory effects of a machine learned theory. 
\vspace{-10pt}

%% file: 02-related.tex
\section{Related work}
\label{related} 

This section summarises related research of game learning and familiarises the reader with the core motivations for our work. We first present a short overview of related investigations in explanatory machine learning of games. Subsequently, we cover various approaches for teaching and learning between humans and machines.
\vspace{-10pt}

\subsection{Explanatory machine learning of games} 

Early approaches to learning game strategies \cite{shapiro1982,quinlan1983} used the decision tree learner ID3 to classify minimax depth-of-win for positions in chess end games. These approaches used carefully selected board attributes as features. However, chess experts had difficulty understanding the learned decision tree due to its high complexity \cite{michie1983}. Methods for simplifying decision trees without compromising their accuracy have been investigated \cite{quinlan1987} on the basis that simpler models are more comprehensible to humans. An early Inductive Logic Programming (ILP) \cite{ILP1991} approach learned optimal chess endgame strategies at depth 0 or 1 \cite{chess1995}. An informal complexity constraint was applied which limits the number of clauses used in any predicate definition to $7\pm2$ clauses. This number is based on the hypothesised limit on human short term memory capacity of $7\pm2$ chunks \cite{Miller1956}. A different approach involving the augmentation of training data with high-level annotations was explored in \cite{Hind2019}. Initialisation requires explanations to be provided for the target data set and the predicative accuracy of explanations is evaluated similarly to the predicative accuracy of labels. 

The earliest reinforcement learning system $MENACE$ (Matchbox Educable Noughts And Crosses Engine) \cite{michie1963} was specifically designed to learn an optimal agent policy for Noughts and Crosses. Later, Q-Learning \cite{watkins1989} and Deep Reinforcement Learning were spawned and have led to a variety of applications including the Atari 2600 games \cite{mnih2015} and the game of Go \cite{Silver2016}. While these systems defeated the strongest human players, they lack the ability to explain the encoded knowledge to humans. Recent approaches such as \cite{zahavy2016} have aimed to explain the policies learned by these models, but the learned strategy is implicitly encoded into the continuous parameters of the policy function which makes their operation opaque to humans. Relational Reinforcement Learning \cite{dvzeroski2001} and Deep Relational Reinforcement Learning \cite{Zambaldi2019DeepRL} have attempted to address these drawbacks by incorporating the use of relational biases to enhance human understandability. Alternatively, case-based policy summary can be provided based on sets of carefully selected states of an agent as representatives of a larger state space to allow humans to gain a limited understanding in short time \cite{summarisation_2019}. 

In \cite{miller2017explanation,MILLER2017}, the author provided a survey of most relevant work in explainable AI and argued that explanatory functionalities were mostly subjective to the developer's view. However, there is a general lack of demonstration on explanatory effect which should be examined by empirical trials and no existing framework accounts for the explanatory harmfulness of machine learned models.  In the context of game playing, we propose a theoretical framework with support of empirical results to characterise helpfulness and harmfulness of machine learning on human comprehension. 

\subsection{Explanations for human problem solving and sequential decision making}
Human problem solving relies on varying degrees of implicit and explicit knowledge -- that is system 1 and system 2 \cite{kahneman2011thinking} --  depending on the problem domain and occasionally on experience of a person \cite{dienes1999theory}. Implicit knowledge which is not available for inspection and verbalisation, is acquired by practice and highly automated \cite{newell1981mechanisms}. In contrast, explicit knowledge, alternatively named declarative knowledge, is inspectable and can be communicated to others \cite{chi2005complex}. For cognitive puzzles such as Tower of Hanoi, it has been shown that parts of the problem solving skills are represented in an explicit way in the form of rules \cite{seger1994implicit}. Communication of problem solving knowledge can be realised in the form of explanations. However, it has been demonstrated in several psychological studies that learners often cannot profit from verbal information when the specific problem solving context is not available to the learners \cite{anderson1997role,berry1995implicit}. However, for intelligent tutoring, it has been suggested that explanations in the form of rules as well as of examples can support learning when given in a specific task context \cite{reed1991use}. Furthermore, it has been shown that learning by doing in combination with explicit verbalisation in the form of explanations is a highly effective learning strategy for cognitive tasks \cite{aleven2002effective}.

One can assume that requirements for explanations to be helpful are different for one-shot classification problems and sequential decision making problems. Explaining the classification decision of a learned model usually refers to the specific instance that is being classified. For example, explanation provided by an intelligent system for identifying the presence of a specific tumor given the image of a tissue sample may include a visual demonstration of the tumor specific tissue and textual information about the size and the position of the tumor in relation to other types of tissue \cite{schmid2020mutual}. In contrast, explaining the decision for a specific action in sequential decision making has to take into account not only the effect of this decision on the current state but also its possible effect on future states \cite{barto1989learning}. Sequential decision making is typical for puzzles such as Tower of Hanoi and for single-person as well as multi-person games. Currently, the function of explanation in games is mostly studied in the context of deep reinforcement learning for Arcade games. One approach is to visualise an agent's current state and factors which affect the agent's decision making \cite{iyer2018transparency}. An exception is a method which summarizes an agent's strategy in a video \cite{sequeira2020interestingness}. In this work, agents do not play optimally and the videos are used to allow the human to assess the capabilities of the agent. For the Ms. Pacman game, it has been demonstrated that visual highlighting can be combined with textual explanations \cite{wang2019verbal}. Studies were pointed out in \cite{stumpf2016} to emphasise a trustworthiness issue of intelligent systems that user's decision making may over-rely on explanatory information provided by intelligent systems even when systems are inaccurate or inappropriate. However, to our knowledge, it has not yet been investigated in what way human comprehension of the agent's behavior profits from multi-modal explanations.

\subsection{Two-way learning between human and machine}
As an emerging sub-field of AI, Machine Teaching \cite{GOLDMAN199520} provides an algorithmic model for quantifying the teaching effort and a framework for identifying an optimized teaching set of examples to allow maximum learning efficiency for the learner. The learner is usually a machine learning model of a human in a hypothesised setting. In education, machine teaching has been applied to devise intelligent tutoring systems to select examples for teaching \cite{Zhu2015,Rafferty2016}. On the other hand, rule-based logic theories are important mechanisms of explanation. Rule-based knowledge representations are generalised means of concept encoding and have a structure analogous to human conception. Mechanisms of logical reasoning, induction and abduction, have long been shown to be highly related to human concept attainment and information processing \cite{Lemke1967,Hobbs2008}. Additionally, humans' ability to apply recursion plays a key role in understanding of relational concepts and semantics of language \cite{Hauser2002} which are important for communication.
 
The process of reconstructing implicit target knowledge which is easy to operate but difficult to describe via machine learning has been explored under the topic of Behavioural Cloning. The cloning of human operation sequence has been applied in various domains such as piloting \cite{Michie1992BuildingSR} and crane operation \cite{Urbancic1994b}. The cloned human knowledge and experience are more dependable and less error-prone due to perceptual and executional inconsistency being averaged across the original behavioural trace. To our knowledge, no existing work has attempted to estimate human errors and target these mistakes in interactive teaching sessions for achieving a measurable "clean up" effect \cite{Michie_cognitive_models} from machine explanations.
\vspace{-5pt}

%% file: 03-theoryframework.tex
\section{Theoretical framework}
\label{framework}
\subsection{Meta-interpretive learning of simple games}
\begin{table}[t]
	\centering
		\caption{A set of win rules is learned by $MIGO$. \textit{MIGO}'s background knowledge contains a general move generator \textit{move/2} and a won classifier \textit{won/1} to encode the minimum rules of the game. The program is dyadic and $win\_1/2$ can be reduced to $win\_1(A,B):-move(A,B),won(B)$ by removing literals after unfolding. A more detailed description of the program learned by $MIGO$ was given in \cite{muggleton2019}.}
	\begin{tabular}{|c|c|}
		\hline 
		\textbf{Depth} & \textbf{Rules}\\ \hline
		1&  \verb+win_1(A,B):- win_1_1_1(A,B),won(B).+\\ 
		&  \verb+ win_1_1_1(A,B):-move(A,B),won(B).+\\ \hline
		2 &  \verb+win_2(A,B):-win_2_1_1(A,B),not(win_2_1_1(B,C)).+\\
		&  \verb+win_2_1_1(A,B):-move(A,B), not(win_1(B,C)).+\\ \hline
		3 & \verb+win_3(A,B):-win_3_1_1(A,B),not(win_3_1_1(B,C)).+\\
		&  \verb+win_3_1_1(A,B):-win_2_1_1(A,B), not(win_2(B,C)).+\\
		\hline
	\end{tabular}
	\label{table:MIGO_rules}
	\vspace{-4pt}
\end{table}

\begin{figure}[t]
	\centering
	\begin{tabular}[c]{cc}
		\minipage{0.3\textwidth}
			\centering
			\includegraphics[width=0.45\textwidth]{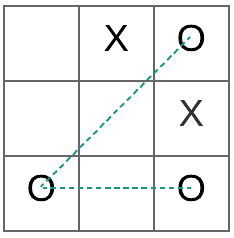}
			\caption{O has two pairs represented in green and X has no pairs.}
			\label{fig:double_lines}
		\endminipage&
		\minipage{0.60\textwidth}
			\centering
			\begin{tabular}{|c|}
				\hline 
				\textbf{Meta-rule}\\ \hline
				$P(A,B) \leftarrow Q(A,B),R(B).$\\
				$P(A) \leftarrow Q(A,B),R(B).$\\
				$P(A) \leftarrow Q(A,S,T),R(A).$\\
				$P(A) \leftarrow Q(A,S,T),R(A,U,V).$\\\hline
			\end{tabular}
			\caption{Letters P, Q, R, S, T, U, V denote existentially quantified second-order variables and A, B, C are universally quantified first-order variables.}
			\label{fig:metarules}
		\endminipage			
	\end{tabular}
	\vspace{-15pt}
\end{figure}

ILP \cite{ILP1991} is a form of machine learning that uses logic programming to represent examples and the background knowledge. The learner aims to induce a hypothesis as a logic program which, together with the background knowledge, entails all of the positive examples and none of the negative examples. Meta-Interpretive Learning (MIL) \cite{muggleton2013,muggleton2014} is a sub-field of ILP which supports predicate invention, dependent learning  \cite{lin2014}, learning of recursions and higher-order programs. Given an input $(\mathcal{B, M, E+, E-})$ where the background knowledge $\mathcal{B}$ is a first-order logic program, meta-rules
$\mathcal{M}$ are second-order clauses, positive examples $\mathcal{E+}$ and negative examples $\mathcal{E-}$ are ground atoms, a MIL algorithm returns a logic program hypothesis $\mathcal{H}$ such that $\mathcal{M} \, \cup \mathcal{H} \, \cup \, \mathcal{B} \models \mathcal{E+}$ and $\mathcal{M} \, \cup \mathcal{H} \, \cup \, \mathcal{B} \not\models \mathcal{E-}$. The background knowledge $\mathcal{B}$ contains primitives which are definitions of concepts represented in the form of predicates. The meta-rules (for examples see Figure \ref{fig:metarules}) contain existentially quantified second-order variables and universally quantified first-order variables. They clarify the declarative bias employed for substitutions of second-order Skolem constants. The resulting first-order theories are thus strictly logical generalisation of the meta-rules.

The MIL game learning framework \textit{MIGO} \cite{muggleton2019} is a purely symbolic system based on the adapted Prolog meta-interpreter Metagol \cite{cropper2016}. \textit{MIGO} learns exclusively from positive examples by playing against the optimal opponent.  For Noughts and Crosses and Hexapawn, \textit{MIGO} learns a rule-like symbolic game strategy (Table \ref{table:MIGO_rules}) that supports human understanding and was demonstrated to converge using less training data compared to Deep and classical Q-Learning. MIGO is provided with a set of three relational primitives, move/2, won/1, drawn/1 which are a move generator, a won and a drawn classifier respectively. These primitives represent the minimal information which a human would know before playing Noughts and Crosses and Hexapawn. For successive values of k, MIGO learns a series of inter-related definitions for predicates $win \_k/2$ for playing as either X or O. These predicates define maintenance of minimax win in k-ply. 

We introduce $MIPlain$\footnote{MIPlain source is available at https://github.com/LAi1997/MIPlain.}, a variant of $MIGO$ which focuses on learning the task of winning for the game of Noughts and Crosses. In addition to learning from positive examples, $MIPlain$ identifies moves which are negative examples for the task of winning. When a game is drawn or lost for the learner, the corresponding path in the game tree is saved for later backtracking following the most updated strategy. $MIPlain$ performs a selection of hypotheses based on the efficiency of hypothesised programs using $Metaopt$ \cite{cropper2019}. 

\begin{table}[t]
	\centering
		\caption{The logic program learned by $MIPlain$ represents a strategy for the first player to win at different depths of the game. The predicate $win\_3\_4/1$ can be reduced to $win\_3\_4(A):-win\_2(A,B)$ by removing literals after unfolding. The program learned by $MIPlain$ can be described as: a board A is won at depth 1 if there exists a move from A to B such that B is won; a board A is won at depth 2 if there exists a move from A to B such that X has exactly two pairs and O has no pairs in B; a board A is won at depth 3 if there exists a move from A to B such that X has exactly one pair in B and there exists a move from B to C such that X does not have any pair in C and C is won at depth 2 for X.} \label{table:rules}
	\begin{tabular}{|c|c|}
		\hline 
		\textbf{Depth} & \textbf{Rules}\\ \hline
		1&  \verb+win_1(A,B):-move(A,B),won(B).+\\ \hline
		2 &  \verb+win_2(A,B):-move(A,B),win_2_1(B).+\\
		&  \verb+win_2_1(A):-number_of_pairs(A,x,2), number_of_pairs(A,o,0).+\\ \hline
		3 & \verb+win_3(A,B):-move(A,B),win_3_1(B).+\\
		&  \verb+win_3_1(A):-number_of_pairs(A,x,1),win_3_2(A).+\\
		&  \verb+win_3_2(A):-move(A,B),win_3_3(B).+\\
		& \verb+win_3_3(A):-number_of_pairs(A,x,0),win_3_4(A).+\\
		& \verb+win_3_4(A):-win_2(A,B),win_2_1(B).+\\ \hline
	\end{tabular}
\end{table}

An additional primitive $number\_of\_pairs/3$ is provided to $MIPlain$ which depicts the number of pairs for a player (X or O) on a given board. A pair is the alignment of two marks of one player, the third square of this line being empty. An example of pairs is shown in Figure \ref{fig:double_lines}. This additional primitive serves as an executional shortcut that reduces the depth of the search when executing the learned strategy. Furthermore, $MIPlain$ is given the meta-rules described in Figure \ref{fig:metarules}, which are two variants of the \textit{postcon} meta-rule with monadic or dyadic head, and two variants of the \textit{conjunction} meta-rule with more than two arguments in either the first or both body literals where existentially quantified argument variables are bound to constants. These meta-rules allow projections of higher dimension predicate definitions onto a monadic setting, therefore enabling the learning of programs with higher-arity predicates. The learned strategy presented in Table \ref{table:rules} describes patterns in game positions in a rule-like manner that the player's optimal move has to satisfy. Due to the instantiation of argument in primitive $number\_of\_pairs/3$, $MIPlain$ learns a program for playing as X assuming X starts the game. For successive values of k, $win \_k/2$ are inter-related predicates which specify status of the game in terms of the number of pairs owned by player X or O and that reflect advantage of player X over player O. 

\vspace{-10pt}
\subsection{Explanatory effectiveness of a machine learned theory}

We extend the machine-aided human comprehension of examples in \cite{US2018} and $C(D, H, E)$ denotes the unaided human comprehension of examples where $D$ is a logic program representing the definition of a target predicate, $H$ is a group of humans and $E$ is a set of examples. Based on the analogy between declarative understanding of a logic program and understanding of a natural language explanation, we describe measures for estimating the degree to which the output of a symbolic machine learning algorithm\footnotemark can be simulated by humans and aid comprehension. \footnotetext{Within the scope of this work, we focus on the symbolic subset of machine learning. However, more general definitions are possible and might be provided by taking into account, for instance, post-hoc interpretations generated from neural networks \cite{schmid2020mutual} and policy summaries extracted from agent-based systems \cite{summarisation_2019}.}

\begin{definition} \textbf{(Machine-explained human comprehension of examples,\\ $C_{ex}(D, H, M(E))$)}: Given a logic program $D$ representing the definition of a target predicate, a group of humans $H$, a theory $M(E)$ learned using machine learning algorithm $M$ and examples $E$, the machine-explained human comprehension of examples $E$ is the mean accuracy with which a human $h \in H$ after brief study of an explanation based on $M(E)$ can classify new material selected from the domain of $D$.
\end{definition}

\begin{definition} \textbf{(Explanatory effect of a machine learned theory, $E_{ex}(D, H, M(E))$)}: Given a logic program $D$ representing the definition of a target predicate, a group of humans $H$, a symbolic machine learning algorithm $M$, the explanatory effect of the theory $M(E)$ learned from examples $E$ is
\begin{gather*}
	E_{ex}(D, H, M(E)) = C_{ex}(D, H, M(E)) - C(D, H, E)
\end{gather*}
\end{definition}

\begin{definition} \textbf{(Beneficial/harmful effect of a machine learned theory)}: Given a logic program $D$ representing the definition of a target predicate, a group of humans $H$, a symbolic machine learning algorithm $M$:
\begin{itemize}
	\item $M(E)$ learned from examples $E$ is \textit{beneficial} to $H$ if $E_{ex}(D, H, M(E)) > 0$
	\item $M(E)$ learned from examples $E$ is \textit{harmful} to $H$ if $E_{ex}(D, H, M(E)) < 0$
	\item Otherwise, $M(E)$ learned from examples $E$ does not have observable effect on $H$
\end{itemize}  
\end{definition}

Within the scope of this work, we relate the explanatory effectiveness of a theory to performance which means that a harmful explanation provided by the machine degrades comprehension of the task and therefore reduces performance. 

\vspace{-10pt}
\subsection{Cognitive window of a machine learned theory}
\label{cognitive_window}
In this section, we suggest a window of a machine learned theory that constraints its explanatory effectiveness. A basic assumption of cognitive psychology and artificial intelligence is that human information processing can be modelled in analogy to symbol manipulation of computers -- respectively its formal characterisation of a Turing Machine \cite{Miller1956,Johnson_laird1986,Newell1990}. More specifically, computational models of cognition share the view that intelligent action is based on manipulation of representations in working memory. In consequence, human inferential reasoning is limited by working memory capacity which corresponds to limitations of tape length and instruction complexity in Turing Machines. 

Besides general restrictions of human information processing, performance can be influenced by internal or environmental disruptions such that the given competencies of a human in a specific domain are not always reflected in observable actions \cite{Chomsky1965,Shohamy1996}. However, it can be assumed that humans -- at least in domains of higher cognition -- are able to explain their actions by verbalising the rules which they applied to produce a given result \cite{Schmid11}. Although rules in general can be classified as procedural knowledge, the ability to verbalise rules makes them part of declarative memory \cite{Anderson1993,Schmid11}. For complex domains, the rules which govern action generation will typically be computationally complex as measured by the Kolmogorov complexity \cite{Kolmogorov1963}. One can assume that increase in complexity can have a negative effect on performance. 

In language processing and in general problem solving, hierarchisation of complex action sequences can make information processing more efficient. Typically, a general goal is broken down into sub-goals as it has been proposed in production system models \cite{Newell1990} as well as in the context of analogical problem solving \cite{Carbonell1985}. Rules which guide problem solving behaviour, for instance in puzzles such as Tower of Hanoi or games such as Noughts and Crosses, might be learned. From a declarative perspective, such learned rules correspond to explicit representations of a concept such as the win-in-two-steps move introduced above. 

Studies of rule-based concept acquisition suggest that human concept learning can be characterised as search in a pool of possible hypotheses which are explored in some order of preference \cite{Bruner1956}. This observation relates to the concept of version space learning introduced in machine learning \cite{Mitchell1982}.  Therefore, for the purpose of experimentation in a noise-free setting, we assume that a) human learners are version space learners with limited hypothesis space search capability and that they use meta-rules to learn sub-goal structure and primitives as background knowledge. We also assume that b) rules can be represented explicitly in a declarative, verbalisable form. Finally, we postulate the existence of a cognitive window such that a machine learned theory can be an effective explanation if it satisfies two constraints: 1) a hypothesised human learning procedure which has a limited search space and 2) a knowledge application model based on the Kolmogorov complexity \cite{Kolmogorov1963}. For the following definitions, we restrict ourselves to learning datalog programs which may take predicates as arguments for representing different data structures but do not include function symbols.

\begin{conjecture} 
	\textbf{(Cognitive bound on the hypothesis space size, $B(P, H)$)}: Consider a symbolic machine learned datalog program $P$ using $p$ predicate symbols and $m$ meta-rules each having at most $j$ body literals. Given a group of humans $H$, $B(P,H)$ is a population-dependent bound on the size of hypothesis space such that at most $n$ clauses in $P$ can be comprehended by all humans in $H$ and $B(P,H)=m^n p^{(1+j)n}$ based on the MIL complexity analysis from \cite{lin2014,cropper2017}.
	\label{proposition:1}
\end{conjecture}

When learned knowledge is cognitively challenging, execution overflows human working memory and instruction stack. We then expect decision making to be more error prone and the task performance of human learners to be less dependable. To account for the cognitive complexity of applying a machine learned theory, we define the cognitive resource of a logic term and atom.

\begin{definition} \textbf{(Cognitive cost of a logic term and atom, $C(T)$)}: Given $T$ a logic term or atom, the cost of $C(T)$ can be computed as follows:
\begin{itemize}
	\item $C(\top) = C(\bot) = 1$
	\item A variable $V$ has cost $C(V) = 1$ 
	\item A constant $c$ has cost $C(c)$ which is the number of digits and characters in $c$
	\item An atom $Q(T_1,T_2,...)$ has cost $C(Q(T_1,T_2,...)) = 1 + C(T_1) + C(T_2) +$ … 
\end{itemize}
\label{def:term_cost}
\end{definition}

\begin{example}
	The Noughts and Crosses position in Figure \ref{fig:double_lines} is represented by the atom $b(e,x,o,e,e,x,o,e,o)$, where b is a predicate representing a board, e is an empty field, o and x are marks on the board. It has cognitive cost $C(b(e,x,o,e,e,x,o,e,o)) = 10$.
\end{example}

Note that we compute cognitive costs of programs without redundancy since repeated literals in programs learned by $MIGO$ and $MIPlain$ were removed after unfolding for generating explanations which are presented to human populations. Also, a game position can be represented by different data types. We ignore the cost due to implementation and only count digits and marks. 

\begin{example}
	An atom $win\_2(b(e,x,o,e,e,x,o,e,o), X)$ with variable $X$ has a cognitive cost $C(win\_2(b(e,x,o,e,e,x,o,e,o), X)) = 12$.
\end{example}

\begin{example}
	A primitive $move(S1, S2)$ which is an atom with variables $S1$ and $S2$ has a cognitive cost $C(move(S1, S2)) = 3$.
\end{example}

We model the inferential process of evaluating training and testing examples as querying a database of datalog programs. The evaluation of a query represents a mental application of a piece of knowledge given a training or testing example. 
The cost of evaluating a query is estimated based on run-time execution stack of a datalog program. In this work, we neglect the cost of computing the sub-goals of a primitive and compute its cost as if it were a normal predicate for simplicity. 

\begin{definition} \textbf{(Execution stack of a datalog program, $S(P, q)$)}: Given a query $q$, the execution stack $S(P, q)$ of a datalog program $P$ is a finite set of atoms or terms evaluated during the execution of $P$ to compute an answer for $q$. An evaluation in which an answer to the query is found ends with value $\top$, and an evaluation in which no answer to the query is found ends with $\bot$. 
\label{def:stack}
\end{definition}

\begin{definition} \textbf{(Cognitive cost of a datalog program, $Cog(P, q)$)}: Given a query $q$, and let $St$ represent $S(P, q)$, the cognitive cost of a datalog program $P$ is
	\begin{gather*}
	Cog(P, q) = \sum_{t \in St} C(t)
	\end{gather*}\label{def:program_cost}
	\vspace{-15pt}
\end{definition}

\begin{example}
	The primitive $move/2$ outputs a valid Noughts and Crosses state from a given input game state; the query is $move(b(x,x,o,e,x,e,o,e,o), S)$. 
	\vspace{-10pt}
	\begin{table}[H]
		\centering
		\begin{tabular}{ c | c }
			$S(move/2, move(b(x,x,o,e,x,e,o,e,o), S)$) & $C(T)$ \\ \hline $move(b(x,x,o,e,x,e,o,e,o), S)$ &  12 \\ \hline
			$move(b(x,x,o,e,x,e,o,e,o), b(x,x,o,e,x,e,o,x,o))$ & 21 \\ \hline
			$\top$ & 1 \\ \hline \hline
			$\textbf{Cog(move/2, move(b(x,x,o,e,x,e,o,e,o), S))}$ & \textbf{34}
		\end{tabular}
		\vspace{-15pt}
	\end{table}
\end{example}

The maintenance cost of task goals in working memory affects performance of problem solving \cite{Carpenter1990}. Background knowledge provides key mappings from solutions obtained in other domains or past experience \cite{Anderson1989,Novick1991} and grants shortcuts for the construction of the current solution process. We expect that when knowledge that provides executional shortcuts is comprehended, the efficiency of human problem solving could be improved due to a lower demand for cognitive resource. Contrarily, in the absence of informative knowledge, performance would be limited by human operational error and would not be better than solving the problem directly. To account for the latter case, we define the cognitive cost of a problem solution that requires the minimum amount of information about the task. 

\begin{definition} \textbf{(Minimum primitive solution program, $\bar{M}_{\phi}(E)$)}: Given a set of primitives $\phi$ and examples $E$, a datalog program learned from examples $E$ using a symbolic machine learning algorithm $\bar{M}$ and a set of primitives $\phi' \subseteq\phi$ is a minimum primitive solution program $\bar{M}_{\phi}(E)$ if and only if for all sets of primitives $\phi'' \subseteq\phi$ where $|\phi''| < |\phi'|$ and for all symbolic machine learning algorithm $M'$ using $\phi''$, there exists no machine learned program $M'(E)$ that is consistent with examples $E$. 
\label{def:min_solution}
\end{definition}

Given a machine learning algorithm $M$ using primitives $\phi$ and examples $E$, a minimum primitive solution program $\bar{M}_{\phi}(E)$ is learned by using the smallest subset of $\phi$ such that $\bar{M}_{\phi}(E)$ is consistent with $E$. A minimum primitive solution program is defined to not use more auxiliary knowledge than necessary but does not necessarily have the minimum cognitive cost over all programs learned with examples $E$. 

\begin{remark}
	Given that the training examples of Noughts and Crosses are winnable and $MIPlain$ uses the set of primitives $\phi$ = $\{ move/2, won/1, number\_of\_pairs/3\}$, a minimum primitive solution program is produced by $MIGO$. This is because $MIGO$ uses primitives $\{move/2$, $won/1\}$ which is a strict subset of $\phi$ for making a move and deciding a win when the input is winnable. Primitives $move/2$ and $won/1$ are also the necessary and sufficient primitives to win Noughts and Crosses and no theory can be learned using a subset of $\phi$ with the cardinality of one.
	\label{remark:1}
\end{remark}

\begin{definition} \textbf{(Cognitive cost of a problem solution, $CogP(E, \bar{M}, \phi, q)$)}: Given examples $E$, primitive set $\phi$, a query $q$ and a symbolic machine learning algorithm $\bar{M}$ that learns a minimum primitive solution, the cognitive cost of a problem solution is
\begin{gather*}
CogP(E, \bar{M}, \phi, q) = Cog(\bar{M}_{\phi}(E), q)
\end{gather*}
where $\bar{M}_{\phi}(E)$ is a minimum primitive solution program.
\label{def:solution_cost}
\end{definition}

\begin{remark}
	 The program $P$ learned by $MIPlain$ has less cognitive cost than the one learned by $MIGO$ except for queries concerning $win\_1/2$. Given sufficient examples $E$, $MIGO$'s learning algorithm as $\bar{M}$, primitive set used by $MIPlain$ $\phi$ = $\{ move/2, won/1, number\_of\_pairs/3\}$, based on Definition \ref{def:stack} to \ref{def:solution_cost}, we have $Cog(P, x_1)$ = $CogP(E, \bar{M}, \phi, x_1)$, $Cog(P, x_2)$ < $CogP(E, \bar{M}, \phi, x_2)$ and $Cog(P, x_3)$ < $CogP(E, \bar{M}, \phi, x_3)$ where $x_i = win_i(s_i, V)$ in which $s_i$ represents a position winnable in $i$ moves and $V$ is a variable.
	 \label{remark:2}
\end{remark}

We give a definition of human cognitive window based on theory complexity during knowledge acquisition and theory execution cost during knowledge application. A machine learned theory has 1) a harmful explanatory effect when its hypothesis space size exceeds the cognitive bound and 2) no beneficial explanatory effect if its cognitive cost is not sufficiently lower than the cognitive cost of the problem solution. 

\begin{conjecture}
    \textbf{(Cognitive window of a machine learned theory)}: Given a logic program $D$ representing the definition of a target predicate, a symbolic machine learning algorithm $M$, a symbolic minimum primitive solution learning algorithm $\bar{M}$ and examples $E$, $M(E)$ is a machine learned theory using the primitive set $\phi$ and belongs to a program class with hypothesis space $S$. For a group of humans $H$,  $E_{ex}$ satisfies both
    
    \begin{enumerate}
        \item $E_{ex}(D, H, M(E))$ < $0$ if $|S|$ > $B(M(E), H)$ 
        \item $E_{ex}(D, H, M(E)) \, \leq \, 0$ if $Cog(M(E), x) \, \geq \, CogP(E, \bar{M}, \phi, x)$ for queries $x$ that $h \in H$ have to perform after study
    \end{enumerate}
\end{conjecture}

We use the defined variant of Kolmogorov complexity as a measure to approximate cognitive cost of applying sequential actions which  does not take empirical data as input. In the following sections \ref{exp_framework} and \ref{experiments}, we concentrate on collecting empirical evidence to support the existence of a cognitive window with bounds (1) and (2) on the explanatory effect. 
\vspace{-15pt}

%% file: 04-expframework.tex
\section{Experimental framework}
\label{exp_framework}

In this section, we describe an experimental framework for assessing the impact of cognitive window on the explanatory effects of a machine learned theory. Our experimental framework involves 1) a set of criteria for evaluating the participants' learning quality from their own textual descriptions of learned strategies and 2) an outline of experimental hypotheses. For game playing, we assume humans are able to explain actions by verbalising procedural rules of strategy. We expect textual answers to provide insights about human decision making and knowledge acquisition. The quality of textual answers can be affected by multiple factors such as motivation, familiarity with the introduced concepts and understanding of the game rules. We take into account these factors in the evaluation criteria. 

\begin{definition} \textbf{(Primitive coverage of a textual answer)}: A textual answer correctly describes a primitive if the semantic meaning of the primitive is unambiguously stated in the response. The primitive coverage is the number of primitives in a symbolic machine learned theory that are described correctly in a textual answer. 
	\label{def:primitive_coverage}
\end{definition}

\begin{definition} \textbf{(Quality of a textual answer, $Q(r)$)}: A textual answer $r$ is checked against the specifications from Table \ref{table:verbal_quality_examples} in an increasing order from criteria level 1 to level 4. $Q(r)$ is the highest level $i$ that $r$ can satisfy. When a response does not satisfy any of the higher levels, the quality of this response is the lowest level 0. \end{definition}

\begin{table}[t]
	\centering
	\caption{Criteria for evaluating textual answers and examples for category $win\_2/2$.}
	\begin{tabular}{ p{1cm} p{5.3cm} p{5cm}  }
		
		$Q(r)$ & Criteria & Exemplary $r$\\ \hline 
		\\ [-0.7em]
		Level 0 & $r$ does not fit into any of the categories below & “Follow the instructions.”\\
		[0.1em]
		Level 1 & One or more primitives in the machine learned theory, directly or by synonyms, are described correctly in $r$ & “This move gives me a pair.” \\
		[0.1em]
		Level 2 & All primitives in the machine learned theory, directly or by synonyms, are described correctly in r & “I should have picked this move to prevent the opponent and get two attacks.”\\
		[0.1em]
		Level 3 & $r$ is unambiguous and all primitives are described correctly, directly or by synonyms, in the same order as in the executional stack of the machine learned theory & "This move gives me two attacks and prevents the opponent from getting a pair." \\
		[0.1em]
		Level 4 & $r$ explains one or more primitives in the machine learned theory in correct causal relations, directly or by synonyms & “This is a good move because by making two pairs and blocking the opponent, the opponent cannot win in one turn and can only block one of my pairs.”\\
		[0.1em]
		\hline
	\end{tabular}
	\label{table:verbal_quality_examples}
	\vspace{-10pt}
\end{table}

To illustrate, we consider the predicate $win\_2/2$ learned by $MIPlain$ (Table \ref{table:rules}). Primitive predicates are $move/2$ and $number\_of\_pairs/3$. We present in Table \ref{table:verbal_quality_examples} a number of examples of textual answers. A high quality response reflects a high motivation and good understanding of game concepts and strategy. On the other hand, a poor quality response demonstrates a lack of motivation or poor understanding.

\begin{definition} \textbf{(High $(HQ)$ / low $(LQ)$ quality textual answer)}: A $HQ$ response $rh$ has $Q(rh) \, \geq \, 3$ and a $LQ$ response $rl$ has $Q(rl) < 3$. \end{definition}

We define the following null hypotheses to be tested in Section \ref{experiments} and describe the motivations. Let $M$ denote a symbolic machine learning algorithm. $E$ stands for examples, $D$ is a logic program representing the definition of a target predicate, $H$ is a group of participants sampled from a human population. $M(E)$ denotes a machine learned theory which belongs to a definite clause program class with hypothesis space $S$. $\bar{M}$ denotes a minimum primitive solution learning algorithm. First, we are interested in demonstrating whether 1) the textual answer quality of learned knowledge reflects comprehension, 2) there exist cognitive bounds for humans to provide textual answers of higher quality and 3) the machine learned theory helps improve the quality of textual answers. \\ [-0.7em]

\noindent \textbf{H1}: \textit{Unaided human comprehension $C(D,H,E)$ and machine-explained human comprehension $C_{ex}(D,H,M(E))$ manifest in textual answer quality $Q(r)$}. 
\hangindent=20pt We examine if high post-training accuracy correlates with high response quality and high primitive coverage of each question category.

\noindent \textbf{H2}: \textit{Difficulty for human participants to provide textual answer increases with quality Q(r)}. 
\hangindent=20pt We examine if the proportion of textual answers reduces with respect to high response quality and high primitive coverage of each question category.

\noindent \textbf{H3}: \textit{Machine learned theory $M(E)$ improves textual answer quality $Q(r)$}. 
\hangindent=20pt We examine if machine-aided learning results in more HQ responses. \\ [-0.7em]

The impact of a cognitive window on explanatory effects is tested via the following hypotheses. $\phi$ is a set of primitives introduced to $H$. Let $x$ denote the set of questions that human $h \in H$ answers after learning.\\ [-0.7em]

\noindent \textbf{H4}: \textit{Learning a complex theory ($|S|$ > $B(M(E), H)$) exceeding the cognitive bound leads to a harmful explanatory effect ($E_{ex}(D, H, M(E))$ < $0$)}. 
\hangindent=20pt We examine if the post-training accuracy, after studying a machine learned theory that participants cannot recall fully, is worse than the accuracy following self-learning. 

\noindent \textbf{H5}: \textit{Applying a theory without a low cognitive cost $(Cog(M(E), x) \, \geq \, CogP(E, \bar{M}, \phi, x))$ does not lead to a beneficial explanatory effect ($E_{ex}(D, H, M(E)) \, \leq \, 0$)}.
\hangindent=20pt We examine if the post-training accuracy, after studying a machine learned theory that is cognitively costly, is equal to or worse than the accuracy following self-learning.
\vspace{-10pt}

%% file: 05-experiments.tex
\section{Experiments}
\label{experiments}
This section introduces the materials and experimental procedure which we designed to examine the explanatory effects of a machine learned theory on human learners. Afterwards, we describe the experiment interface and present experimental results.
\vspace{-10pt}

\subsection{Materials}
\label{materials}
We assume that Noughts and Crosses is a widely known game a lot of participants of the experiments are familiar with. This might result in many participants already playing optimally before receiving explanations, leaving no room for potential performance increase. In order to address this issue, the \textit{Island Game} was designed as a problem isomorphic to Noughts and Crosses. \cite{simon1976} define isomorphic problems as "problems whose solutions and moves can be placed in one-to-one relation with the solutions and moves of the given problem". This changes the superficial presentation of a problem without modifying the underlying structure. Several findings imply that this does not impede solving the problem via \textit{analogical inference} if the original problem is consciously recognized as an analogy; on the other hand, the prior step of initially identifying a helpful analogy via \textit{analogical access} is highly influenced by superficial similarity \cite{gentner1985,holyoak1987,reed1990}. Given that the Island Game presents a major re-design of the game surface, we expect that participants will less likely recall prior experience of Noughts and Crosses that would facilitate problem solving, leading to less optimal play initially and more potential for performance increase.

The Island Game (Figure \ref{fig:pretest}) contains three islands, each with three territories on which one or more resources are marked. The winning condition is met when a player controls either all territories on one island or three instances of the same resource. The nine territories resemble the nine fields in Noughts and Crosses and the structure of the original game is maintained in regard to players' turns, possible moves, board states and win conditions. This isomorphism masks a number of spatial relations that represent the membership of a field to a win condition. In this way, the fields can be rearranged in an arbitrary order without changing the structure of the game. 

\vspace{-10pt}
\subsection{Methods and design}

\begin{table}[t]
	\centering
	\caption{Summary of experiment parts. Participants played one mock game against a random computer player for the more difficult Island Game. After selecting a move in training and regardless of its correctness, participants received the labels of the two moves presented; treated participants additionally received explanations generated from $MIPlain$'s learned program. We introduced the primitive set used by $MIPlain$.}
	\label{table:experiment_procedure}
	\begin{tabular}{ p{1.6cm} p{4cm} p{0.3cm} p{4.5cm} }
		Stage & Participant's assignment & No. & Question format\\ \hline 
		\\ [-0.7em]
		Intro & Understand rules to move and win & 1 & practice\\ \hline 
		Pre-training & Choose the optimal move & 15 & five canonical positions each for win\_1, win\_2 \& win\_3\\ \hline 
		Training & Understand the concept of pairs; choose the optimal move and reflect on the choice & 9 & two choices each for win\_1, win\_2 \& win\_3; presentation of the labels\\ \hline 
		Post-training & Choose the optimal move & 15 & five canonical positions each for win\_1, win\_2 \& win\_3; Rotated and flipped from pre-training questions.\\ \hline 
		Open questions & Describe the strategy of a previously made move & 6 & Questions requiring textual answer\\ \hline 
		Survey & Provide gender, age group \& education level & 3 & multiple choices \\ \hline
	\end{tabular}
\end{table}

We use two experiment interfaces, one for Noughts and Crosses and another one for the Island Game. A human player always plays as player one (X for Noughts and Crosses and Blue for the Island Game) and starts the game. For both, we adopt a two-group pre-training post-training design (Table \ref{table:experiment_procedure}). We first introduce to participants rules of the game and the concept of pairs. In the pre-training stage, performance of participants in both self learning and machine-aided learning groups are measured in an identical way. During training, they are able to see correct answers of some game positions. In the post-training, performance of both self-learning and machine-aided groups are evaluated in the exact same way as in the pre-training. This experiment setting allows to evaluate the degree of change in performance as the result of explanations. Each question in pre- and post-training is the presentation of a board for which it is the participant's turn to play. They are asked to select what they consider to be the optimal move. A question category of $win_i$ denotes a game position winnable in $i$ moves of the human player. An exemplary question is shown in the Figure \ref{fig:pretest}. The post-training questions are rotated and flipped from pre-training questions. In each test, only 15 questions are given to limit experiment duration to one hour. The response time of participants was recorded for each pre-training and post-training question.

\begingroup
\setlength{\intextsep}{0pt}
\setlength{\columnsep}{10pt}
\begin{wrapfigure}[16]{r}{0.65\textwidth}
	\includegraphics[width=0.60\textwidth]{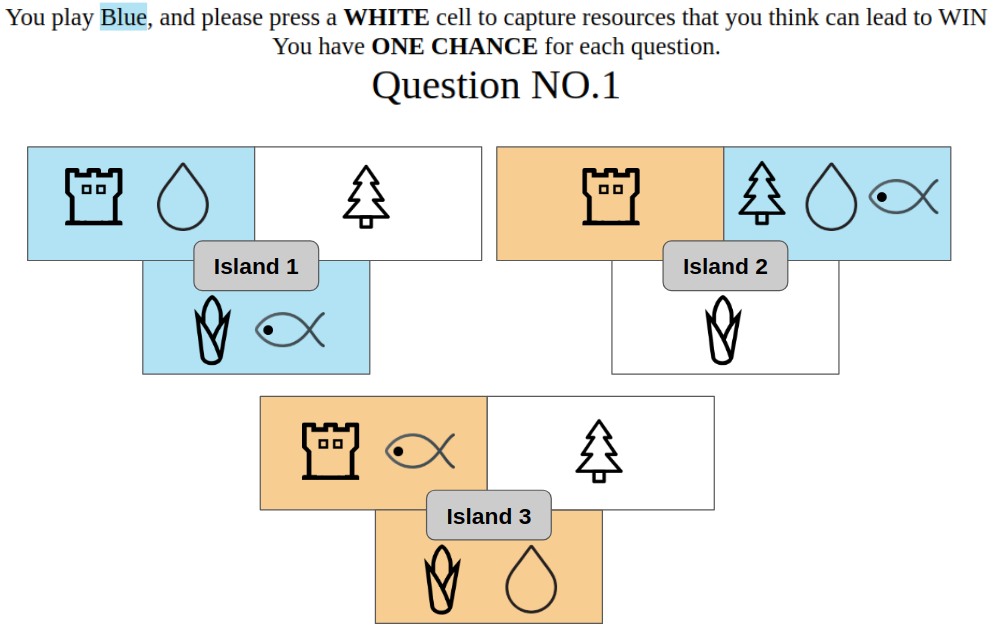}
	\caption{Example of pre- and post-training question for the Island Game. A board is presented to the participant to select a move that he or she thinks is optimal.}
	\label{fig:pretest}
\end{wrapfigure}

The treatment was applied to the machine-aided group. Various studies \cite{aleven2002effective,anderson1997role,berry1995implicit,reed1991use} suggested explanations are most effective for human learning when presented with examples and in a specific task context. Therefore, we have employed textual explanations to verbalise machine learned knowledge for a sequence of game states and these textual explanations are grounded to instantiate game states in the context to provide visualisation of game boards. During treatment, we present both visual and textual explanations in order for participants who are not familiar with the designed game domain to profit the most from explanations. Learned first-order theories have been translated with manual adjustments based on primitives provided to all participants and to $MIPlain$. An exemplary explanation is shown in Figure \ref{fig:explanation}. Both visual and textual explanations preserve the structure of hypotheses without redundancy and account for the reasons that make a move correct (highlighted in green). Contrastive explanations are presented for the possible sequence of wrong moves in participant's turns (highlighted in red) by comparing against $MIPlain$'s learned theory. Conversely, during training, the self-learning group did not receive the treatment and was presented with similar game positions without the corresponding visual and textual explanations. For the Island Game experiments, we recorded an English description of the strategy they used for each of the selected post-training questions. Participants are presented previously submitted answers, one at a time along with a text input box for written answers. Moves for these open questions are selected from post-training with a preference order from wrong and hesitant moves to consistently correct moves. We associate hesitant answers with higher response times. A total of six questions are selected based on individual performance during the post-training. 

\endgroup

\vspace{-10pt}
\subsection{Experiment results} 

We conducted three trial experiments\footnote{\label{note1}Raw data are available upon request from the authors.} using the interface with Noughts and Crosses questions and explanations. These experiments were carried out on three samples: an undergraduate student group from Imperial College London, a junior student group from a German middle school and a mixed background group from Amazon Mechanical Turk\footnote{AMT (www.mturk.com) is an online crowdsourcing platform which we used to recruit experiment participants.} (AMT). No consistent explanatory effects could be observed for any of the mentioned samples. The problem solving strategy that humans apply can be affected by factors such as task familiarity, problem difficulty, and motivation. For instance, \cite{Schmid2000} suggested that a rather superficial analogical transfer of a strategy is applied when a problem is too difficult or when there is no reason to gain a more general understanding of a problem. Given that the majority of subjects achieved reasonable initial performance, we ascribe the reason of such results to experience with the game and complexity of explanations. The game familiarity of adult groups led to less potential for performance improvement. Early middle school students had limited attention and were overwhelmed by information intake. Alternatively, we focused on specially designed experiment materials in the following experiments. 

\subsubsection{Island Game with open questions}
A sample from Amazon Mechanical Turk and a student sample from the University of Bamberg participated in experiments\footnoteref{note1} that used the interface with Island Game questions and explanations. To test hypotheses \textbf{H1} to \textbf{H5}, we employed a quantitative analysis on test performance and a qualitative analysis on textual answers. A sub-sample with a mediocre performance on pre-training questions of all categories within one standard deviation of the mean was selected for the performance analysis. This aims to discount the ceiling effect (initial performance too high) and outliers (e.g. struggling to use the interface). We employed 5\% significance levels for testing experimental results.

\begin{figure}[t]
    \centering
    \begin{tabular}{cc}
        \begin{subfigure}{0.45\textwidth}
	        \includegraphics[width=\linewidth]{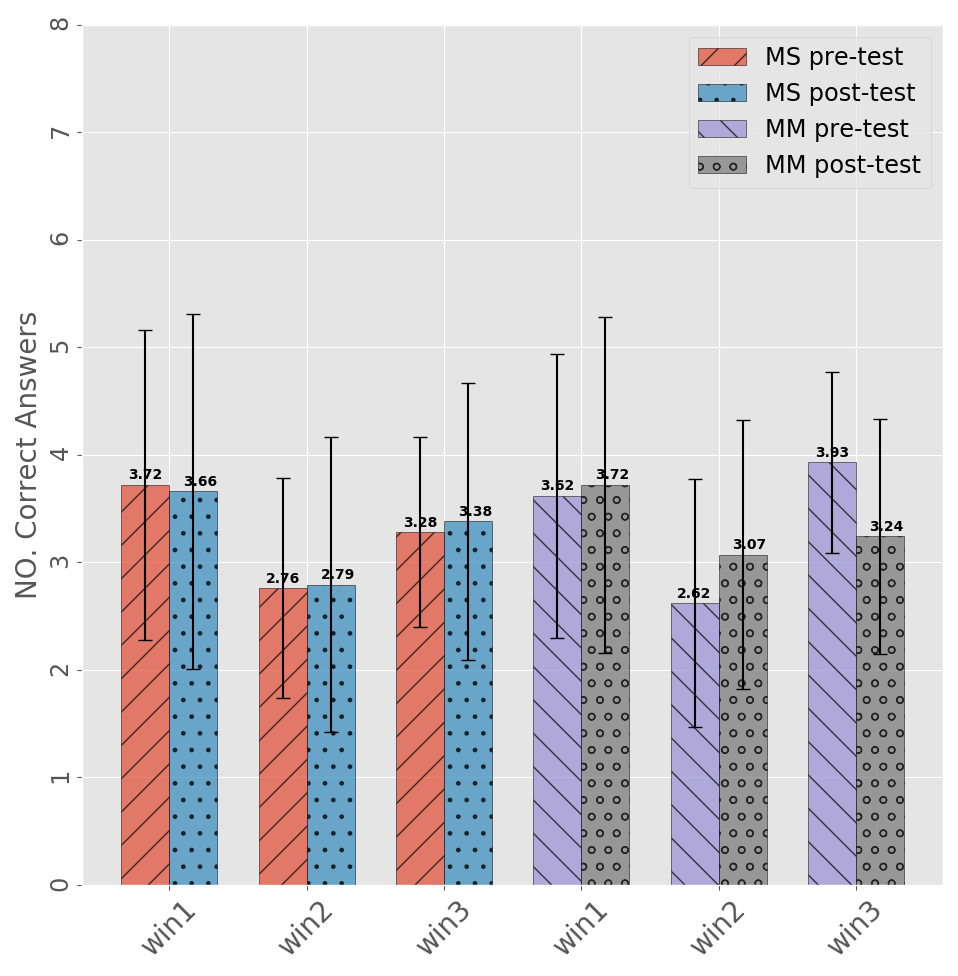}
	        \caption{Mixed background self learning and machine-aided learning.}
	    \label{fig:island_amazon_performance_test}
        \end{subfigure} & 
        \begin{subfigure}{0.45\textwidth}
	        \includegraphics[width=\linewidth]{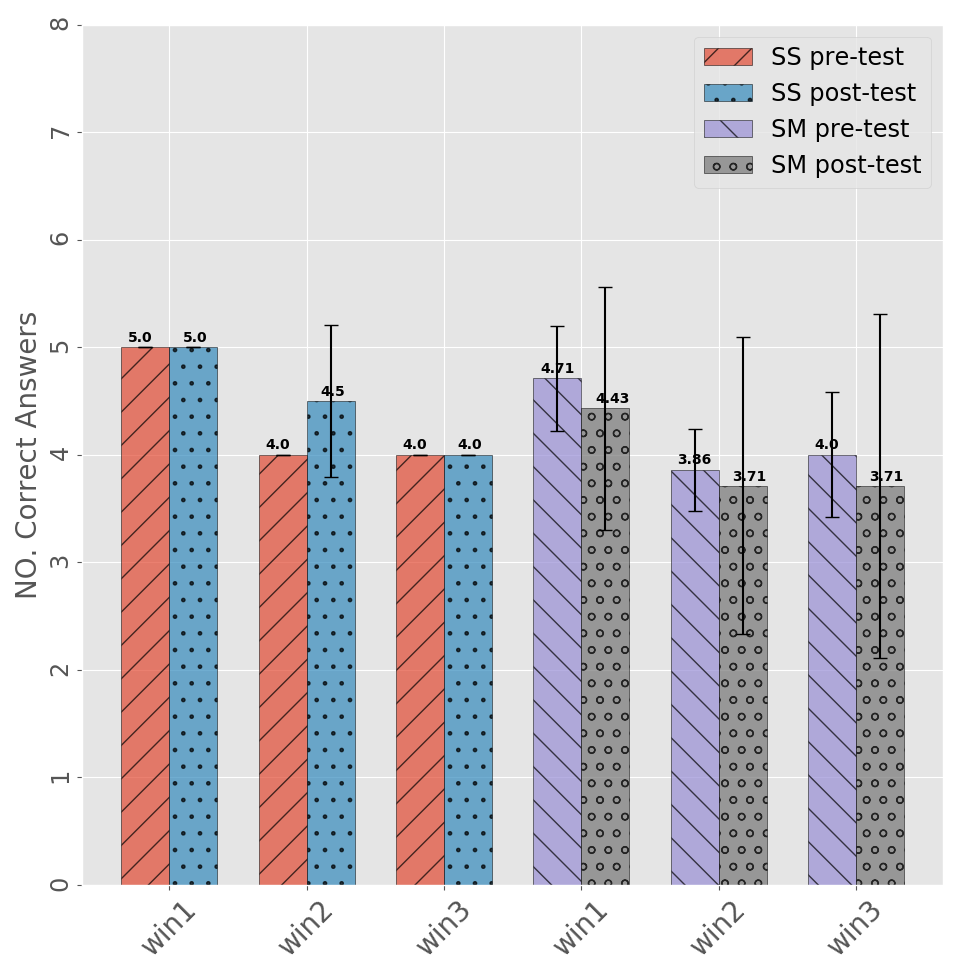}	
	        \caption{Student self learning and machine-aided learning.}
	        \label{fig:island_bamberg_performance_test}
        \end{subfigure}
    \end{tabular}
    \caption{Number of correct answers in pre- and post-training with respect to question categories.}
    \label{fig:performance_test}
    \vspace{-10pt}
\end{figure}

From AMT sample, we had 90 participants who were 18 to above 65 years old. A sub-sample of 58 participants with a mediocre initial performance was randomly partitioned into two groups, \textbf{MS} (Mixed background Self learning $n = 29$) and \textbf{MM} (Mixed background Machine-aided learning, $n = 29$). A different sub-sample of 30 participants completed open questions and was randomly split into two groups, \textbf{MSR} (Mixed background Self learning and strategy Recall, $n = 15$) and \textbf{MMR} (Mixed background Machine-aided learning and strategy Recall, $n = 15$). As shown in Figure \ref{fig:island_amazon_performance_test}, in category $win\_2$, \textbf{MM} post-training had a better comprehension ($p$ = 0.028) than \textbf{MS} post-training while \textbf{MM} and \textbf{MS} had similar pre-training performance in this category. Results in category $win\_2$ indicate that explanations have a beneficial effect on \textbf{MM}. However, \textbf{MM} did not have a better comprehension on $win\_1$ than \textbf{MS} given the same initial performance. In addition, \textbf{MM} had the same initial performance as \textbf{MS} in category $win\_3$ but \textbf{MM}'s performance reduced after receiving explanations of $win\_3$ ($p$ = 0.005).

From a group of students involved in a Cognitive Systems course at the University of Bamberg, we had 13 participants who were 18 to 24 years old and a few outliers between 25 and 54 years. All participants were asked to complete open questions and were randomly split into two groups, \textbf{SSR} (Student Self learning and strategy Recall, $n = 4$) and \textbf{SMR} (Student Machine-aided learning and strategy Recall, $n = 9$). A sub-sample of 9 with a mediocre initial performance was randomly divided into \textbf{SS} (Student Self learning, $n = 2$) and \textbf{SM} (Student Machine-aided learning, $n = 7$). The imbalance in the student sample was caused by a number of participants leaving during the experiment. The machine-aided learning results show large performance variances in post-training as evidence for insignificant levels of performance degradation.

\begin{table}[t]
	\centering
	\caption{The number and accuracy of HQ and LQ responses for groups \textbf{MSR}, \textbf{MMR}, \textbf{SSR}, \textbf{SMR} and each question category. For $win\_3$, the most mentally challenging category of all three, no HQ response was given.}
	\vspace{-5pt}
	\begin{tabular}{ p{1cm} p{4cm} p{1.8cm} p{1.8cm} p{1.8cm}  }
		& & $win\_1$ & $win\_2$ & $win\_3$ \\
		\hline
		MSR & No. HQ / post-train accuracy & 9 / 0.889 & 1 / 1.0 & -\\
		& No. LQ / post-train accuracy & 19 / 0.421 & 32 / 0.406 & 29 / 0.517\\
		\hline
		 MMR & No. HQ / post-train accuracy & 8 / 1.00 & 2 / 1.00 & -\\
		& No. LQ / post-train accuracy & 16 / 0.250 & 35 / 0.486 & 29 / 0.483\\
		\hline
		SSR & No. HQ / post-train accuracy & 6 / 1.00 & 1 / 1.00 & - \\
		& No. LQ / post-train accuracy & 0 / 0.00 & 8 / 0.750 & 9 / 0.667\\
		\hline
		SMR & No. HQ / post-train accuracy & 9 / 1.00 & 9 / 0.778 & - \\
		& No. LQ / post-train accuracy & 3 / 0.00 & 14 / 0.571 & 19 / 0.737\\
		\hline
	\end{tabular}
	\label{table:sanity_test}
	\vspace{-10pt}
\end{table}

In Table \ref{table:sanity_test}, we identified that participants who were able to provide high quality responses for their test answers scored higher on these questions. This is not the case for $win\_3$, however, due to the high difficulty of providing good description of strategy for $win\_3$ category. Additionally, in the $win\_2$ category, both machine-aided groups (\textbf{MMR}: 2/(2+35), \textbf{SMR}: 9/(9+14)) have greater proportions of high quality responses than self learning groups (\textbf{MSR}: 1/(1+32), \textbf{SSR}: 1/(1+8)). Also, we observed a pattern in which there are less HQ responses than LQ responses in $win\_1$ and $win\_2$ categories. This pattern is more significant in $win\_2$ category. 

\begin{figure}
	\centering
	\begin{tabular}[c]{cc}
		\begin{subfigure}[b]{0.45\textwidth}
			\includegraphics[width=\textwidth]{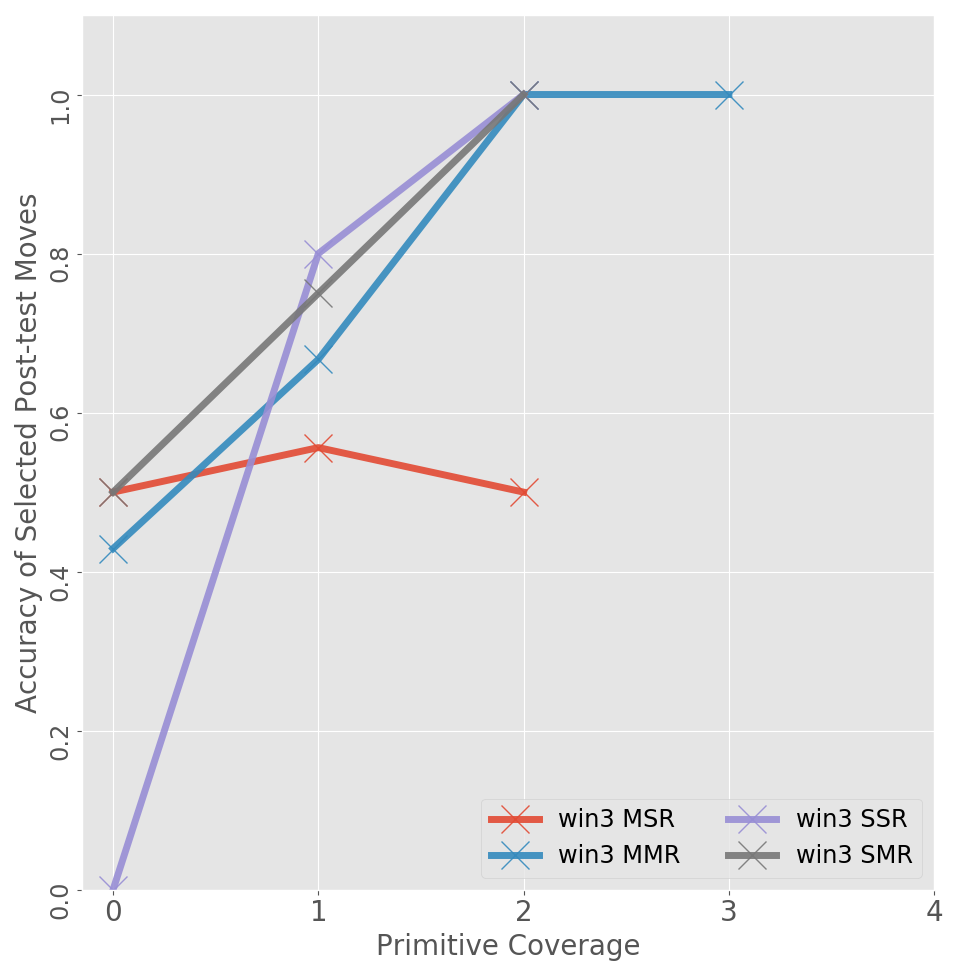}
			\caption{The accuracy of textual answers increases with respect to the number of primitives covered.}
			\label{fig:coverage_accuracy}
		\end{subfigure}&
		\begin{subfigure}[b]{0.45\textwidth}
			\includegraphics[width=\textwidth]{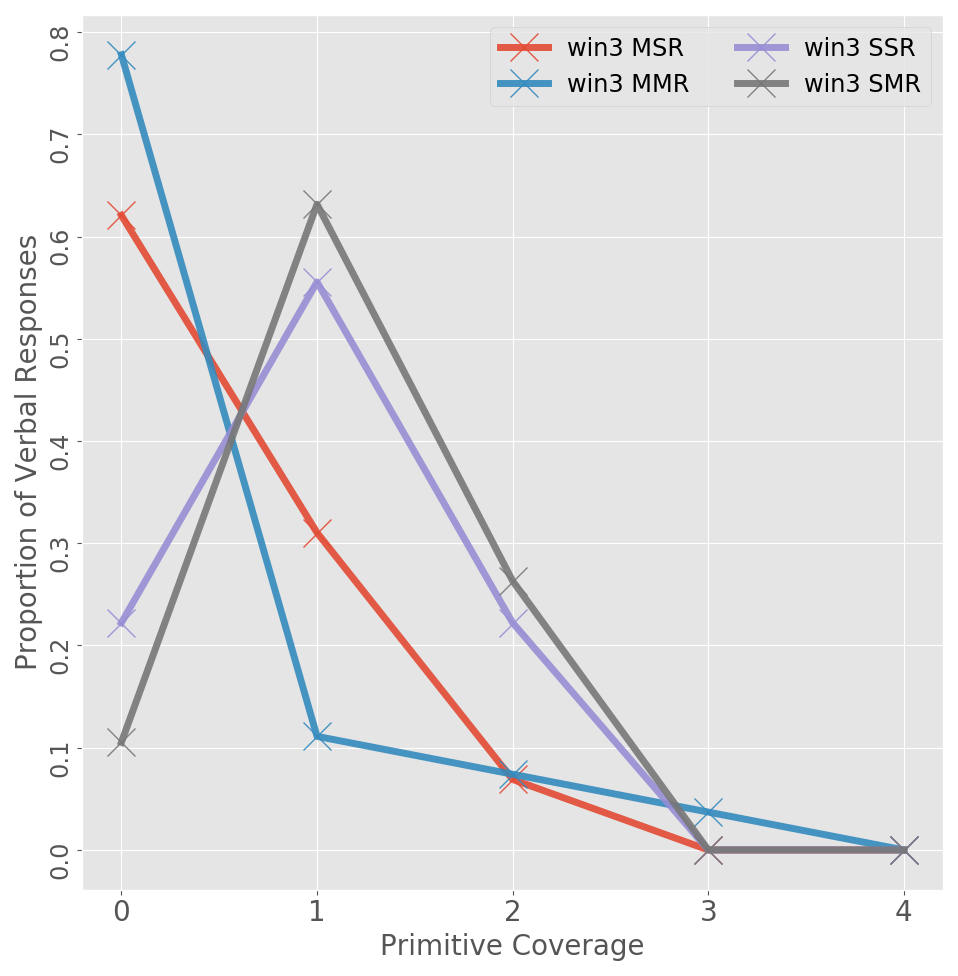}
			\caption{The proportion of quality textual answers decreases with respect to the number of primitives covered.}
			\label{fig:coverage_number_res}
		\end{subfigure}
	\end{tabular}
	\caption{$win\_3$ reuses $win\_2$ and uses four $number\_of\_pairs/3$ when unfolded. In Figure \ref{fig:coverage_number_res}, both mixed background groups (\textbf{MSR} and \textbf{MMR}) had lower proportions of responses covering one predicate than student groups (\textbf{SSR} and \textbf{SMR}). Mixed background and student groups could not provide a significant proportion of response covering more than one and two primitives respectively (Figure \ref{fig:coverage_accuracy}).}
	\label{fig:coverage}
	\vspace{-10pt}
\end{figure}

Figure \ref{fig:coverage} illustrates the difficulty of providing good quality textual answer for the non-trivial category $win\_3$. Since $win\_1$ contains only two predicates, we examined primitive coverage of non-trivial categories $win\_2$ and $win\_3$. However, for clarity of presentation, we only show category $win\_3$ which has more remarkable trends. When counting primitives based on Definition \ref{def:primitive_coverage}, we only consider the constraint $number\_of\_pairs/3$ and ignore the move generator $move/2$ as participants were required to make a move when they answered a question.

In Figure \ref{fig:coverage_accuracy}, we plotted primitive coverage against the accuracy of post-training answers that were selected as open questions. We observed a major \textit{monotonically increasing trend} in accuracy with respect to primitive coverage. This indicates that high matching between textual answers and the machine learned theory correlates with high performance. In Figure \ref{fig:coverage_number_res}, we observed \textit{downward curves} for \textbf{MSR} and \textbf{MMR} in the number of textual answers from the lower to the higher primitive coverage. More responses were provided by \textbf{SSR} and \textbf{SMR} covering \textit{one primitive} than \textbf{MSR} and \textbf{MMR}. Participants gave very few responses that cover \textit{more than two} primitives. Based on the learned theory\footnote{The translation of the learned theory into textual and visual explanations does not contain redundant parts.} of $MIPlain$ in Table \ref{table:rules}, the results suggest an increasing difficulty to provide complete strategy descriptions \textit{beyond two (mixed background groups) and four (student groups) clauses} of $win\_3$. 
\vspace{-10pt}

%% file: 06-discussion.tex
\subsection{Discussion}
\label{discussion}
Results concerning null hypotheses \textbf{H1} to \textbf{H5} are summarised in Table \ref{table:discussion_island1} and \ref{table:discussion_island2}. We assume that (H1 Null) comprehension does not correlate with textual answer quality. To examine this hypothesis, we analyse the results in two steps. First, results of HQ responses in two categories (Table \ref{table:sanity_test}) suggest that being able to provide better textual answers of strategy corresponds to a high comprehension. Second, we examined the coverage of primitives (specifically for LQ responses of $win\_3$) in textual answers (Figure \ref{fig:coverage_accuracy}). Evidence in all categories shows a correlation between comprehension and the degree of textual answer matching with explanations. We reject the null hypothesis in all categories which implies the confirmation of H1.

\begin{table}[t]
	\centering
	\caption{Hypotheses concerning quality of textual answers and comprehension. C stands for \textbf{c}onfirmed, N denotes \textbf{n}ot confirmed, H stands for \textbf{h}ypothesis. Test outcomes are presented for $win\_1$, $win\_2$ and $win\_3$ categories.}
	\begin{tabular}{ p{0.3cm} p{8cm} p{0.6cm} p{0.6cm} p{0.6cm} }
		H & & $win\_1$ & $win\_2$ & $win\_3$\\ \hline 
		\\ [-0.7em]
		H1 & Human comprehensions manifest in textual answer quality & C & C & C\\
		[0.1em]
		H2 & Difficulty for human participants to provide textual answer increases with  textual answer quality & C & C & C\\
		[0.1em]
		H3 & Machine learned theory improves textual answer quality & N & C & N\\
		\hline
	\end{tabular}
	\label{table:discussion_island1}
\end{table}

In addition, we assume that (H2 Null) the difficulty for human participants to provide textual answer is not affected by textual answer quality. Since high response quality is difficult to achieve (Table \ref{table:sanity_test}) and it is challenging to correctly describe all primitives (Figure \ref{fig:coverage_number_res}), we reject this null hypothesis for all categories and confirm H2 as it is increasingly difficult for participants to provide higher quality textual answer. Hence, two additional trends we observed from the same figure suggest two mental barriers of learning. As we assume a human sample is a collection of version space learners, the search space of participants is limited to programs of size two (mixed background groups) and four (student groups). When $H$ is taken as the student sample and $P$ to be the machine learned theory on winning the Island Game, the cognitive bound \textbf{$B(P, H) = m^4 * p^{4(j+1)} = 4^4 * 2^{12}$} corresponds to the hypothesis space size for programs with four clauses (four metarules are used with at most two body literals in each clause, primitives are $move/2$ and $number\_of\_pairs/3$). 

Furthermore, we assume that (H3 Null) machine learned theory does not improve textual answer quality. Results (Table \ref{table:sanity_test}) show higher proportion of HQ responses for machine-aided learning than self-learning in category $win\_2$. Thus, for $win\_2$, we reject this null hypothesis which means H3 is confirmed in category $win\_2$ where the machine explanations result in more high quality textual answers being provided. 

We assume that (H4 Null) learning a descriptively complex theory does not affect comprehension harmfully. When $P$ is the program learned by $MIPlain$, $B(P, H)$ for two samples correspond to program class with size no larger than 4. Only $win\_3$ which has a larger size of seven after unfolding exceeds these cognitive bounds. As harmful effects (Figure \ref{fig:island_amazon_performance_test} and \ref{fig:island_bamberg_performance_test}) have been observed in category $win\_3$, this null hypothesis is rejected and H4 is confirmed as learning a complex machine learned theory has a harmful effect on comprehension. We also assume that (H5 Null) applying a theory without a sufficiently low cognitive cost has a beneficial effect on comprehension. According to Remark \ref{remark:2}, given sufficient training examples $E$, $MIGO$'s learning algorithm as $\bar{M}$ and $\phi$ = $\{ move/2, won/1, number\_of\_pairs/3\}$, the predicate $win\_1$ in $MIPlain$'s learned theory does not have a lower cognitive cost: for all queries x of winning in one move, $Cog(win\_1, x) \, \geq \, CogP(E, \bar{M}, \phi, x)$. We reject this null hypothesis since no significant beneficial effect has been observed in category $win\_1$. Therefore, we confirm H5 -- knowledge application requiring much cognitive resource does not result in better comprehension.

\begin{table}[t]
	\centering
	\caption{Hypotheses concerning cognitive window and explanatory effects. C stands for \textbf{c}onfirmed, H stands for \textbf{h}ypothesis, T stands for \textbf{t}est outcome. }
	\begin{tabular}{ p{0.3cm} p{10.5cm} p{0.3cm} }
		H &  & T \\ \hline 
		\\ [-0.7em]
		H4 & Learning a complex theory exceeding the cognitive bound leads to a harmful explanatory effect & C \\
		[0.1em]
		H5 & Applying a learned theory without a low cognitive cost does not lead to a beneficial explanatory effect & C \\
		\hline
	\end{tabular}
	\label{table:discussion_island2}
	\vspace{-10pt}
\end{table}

The performance analysis (Figure \ref{fig:island_amazon_performance_test}) demonstrates a comprehension difference between self learning and machine-aided learning in category $win\_2$. An explanatory effect has not been observed for the student sample. While the conflicting results suggest that a larger sample size would likely ensure consistency of statistical evidence, the patterns in results suggest more significant results in category $win\_2$ than $win\_1$ and $win\_3$. The predicate $win\_2$ in the program learned by $MIPlain$ satisfies both constraints on hypothesis space bound for knowledge acquisition and cognitive cost for knowledge application. In addition, the cognitive window explains the lack of beneficial effects of predicates $win\_1$ and $win\_3$. The former does not have a lower cognitive cost for execution so that operational errors cannot be reduced, thus there has been no observable effects. The latter is a complex rule with a larger hypothesis space for human participants to search from and harmful effects have been observed due to partial knowledge being learned. 
\vspace{-10pt}

%% file: 07-conclusions.tex
\section{Conclusions and further work}
\label{conclusion}

While the focus of explainable AI approaches has been on explanations of classifications \cite{adadi2018peeking}, we have investigated explanations in the context of game strategy learning. In addition, we have explored both beneficial and harmful sides of the AI's explanatory effect on human comprehension. Our theoretical framework involves a cognitive window to account for the properties of a machine learned theory that lead to improvement or degradation of human performance. The presented empirical studies have shown that explanations are not helpful in general but only if they are of appropriate complexity -- being neither informatively overwhelming nor more cognitively expensive than the solution to a problem itself. It would appear that complex machine learning models and models which cannot provide abstract descriptions of internal decisions are difficult to be explained effectively. However, it remains an open question how one can examine non-explainability. This is an important question since a positive outcome implies the limit of scientific explanations. In this work, a conservative approach has been taken and we have obtained preliminary results from a rather narrow domain. We have acknowledged that participant groups vary greatly in size which might be extended with studies on a broader range of problems with larger samples. Similar metrics that relate to explanatory effects but expand beyond symbolic machine learning have great potentials for future work. The noise-free framework for cognitive window in this work might also be extended with hypotheses that take inconsistency of data into consideration.

To explain a strategy, typically goals or sub-goals must be related to actions which can fulfill these goals. If the strategy involves to keep in mind a stack of open sub-goals -- as for example the Tower of Hanoi \cite{Altmann2002,Schmid11} -- explanations might become more complex than figuring out the action sequence. Based on \cite{Bruner1956}, knowledge is learned by humans in an incremental way, which was recently emphasized by \cite{Zeller2017AHL} on human category learning. Given problems whose solutions can be effectively divided into sufficiently small parts, a potential approach to improve explanatory effectiveness of a machine learned theory is to process complex concepts into smaller chunks by initially providing simple-to-execute and short sub-goal explanations. Mapping input to another sub-goal output thus consumes lower cognitive resources and improvement in performance is more likely. It is worth investigating for future work a teaching procedure involving a sequence of teaching sessions that issues increasingly difficult tasks and explanations. Yet, Abstract descriptions might be generated in the form of invented predicates as it has been shown in previous work on ILP as an approach to USML \cite{US2018}. An example for such an abstract description for the investigated game is the predicate $number\_of\_pairs/3$. Therefore, learning might be organised incrementally, guided by a curriculum \cite{bengio2009curriculum,telle2019teaching}.

\begin{figure}[t]
	\centering
	\begin{tabular}[c]{cc}
		\minipage{0.45\textwidth}
		\centering
		\includegraphics[width=\textwidth]{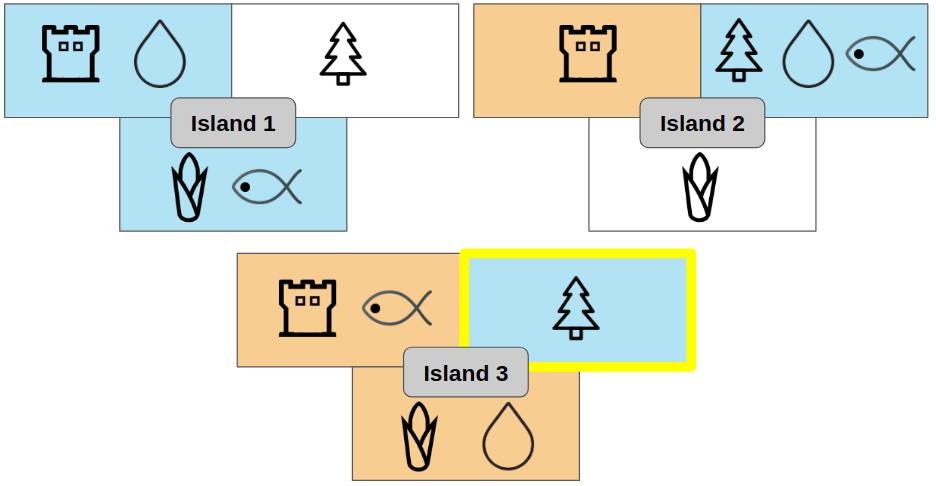}
		\endminipage&
		\minipage{0.40\textwidth}
		\centering
		\begin{verbatim}
		Participant's strategy:
		win_1(A,B):-move(A,B),
		            number_of_pairs(B,o,0).
		Correct strategy:
		win_1(A,B):-move(A,B),
		            won(B).
		\end{verbatim}
		\endminipage			
	\end{tabular}
	\caption{Left: participant's chosen move from the initial position in Figure \ref{fig:pretest}. Right: $Metagol$ one-shot learns from participant's move a program representing his strategy. The learned program represents a strategy to prevent player Orange (who would play O in Noughts and Crosses) from occupying the entire island No.3 rather than going for a full occupancy on island No.1 which is an immediate win and a mismatch between learned and taught knowledge. }
	\vspace{-15pt}
	\label{fig:demo}
\end{figure} 

In addition, the current teaching procedure, which only specifies humans as learners, could be augmented to enable two-way learning between human and machine. Human decisions might be machine learned and explanations would be provided based on estimation of human errors during the course of training. A simple demonstration of this idea is presented in Figure \ref{fig:demo}. We would like to explore, in the future, an interactive procedure in which a machine iteratively re-teaches human learners by targeting human learning errors via specially tailored explanations. \cite{BRATKO1995143} suggested it is crucial for machine produced clones to be able to represent goal-oriented knowledge which is in a form that is similar to human conceptual structure. Hence, MIL is an appropriate candidate for cloning since it is able to iteratively learn complex concepts by inventing sub-goal predicates. We hope to incorporate cloning to predict and target mistakes in human learned knowledge from answers in a sequence of re-training. We expect a "clean up" on operation errors of human behaviours from empirical experiments by presenting appropriate explanations in re-training. Such corrections and improvements guided by identified errors in a human strategy are also helpful in the context of intelligent tutoring \cite{ZellerSchmid16} where classic strategies such as algorithmic debugging \cite{shapiro1982algorithmic} can be applied to make humans and machines learn from each other.
\vspace{-15pt}

%% file: 08-acknowledgements.tex
\section*{Acknowledgements}
The contribution of the authors from University of Bamberg is part of a project funded by the Deutsche Forschungsgemeinschaft (DFG, German Research Foundation) -- 405630557 (PainFaceReader). The second author acknowledges support from the UK's EPSRC Human-Like Computing Network, for which he acts as director. 
\vspace{-10pt}

%% file: main.bbl
\begin{thebibliography}{10}

\bibitem{adadi2018peeking}
A.~Adadi and M.~Berrada.
\newblock Peeking inside the black-box: {A} survey on explainable artificial
  intelligence {(XAI)}.
\newblock {\em IEEE Access}, 6:52138--52160, 2018.

\bibitem{aleven2002effective}
V.~Aleven and K.~R. Koedinger.
\newblock An effective metacognitive strategy: Learning by doing and explaining
  with a computer-based cognitive tutor.
\newblock {\em Cognitive science}, 26(2):147--179, 2002.

\bibitem{Altmann2002}
E.~Altmann and J.~G. Trafton.
\newblock Memory for goals: An activation-based model.
\newblock {\em Cognitive Science}, 26:39--83, 2002.

\bibitem{summarisation_2019}
O.~Amir, F.~Doshi-Velez, and D.~Sarne.
\newblock Summarizing agent strategies.
\newblock {\em Autonomous Agent Multi-Agent System}, 33:628–644, 2019.

\bibitem{anderson1997role}
J.~R. Anderson, J.~M. Fincham, and S.~Douglass.
\newblock The role of examples and rules in the acquisition of a cognitive
  skill.
\newblock {\em Journal of experimental psychology: learning, memory, and
  cognition}, 23(4):932, 1997.

\bibitem{Anderson1993}
J.~R. Anderson, N.~Kushmerick, and C.~Lebiere.
\newblock {\em Rules of the Mind}, chapter The Tower of Hanoi and goal
  structures, pages 121--142.
\newblock Hillsdale, NJ: L. Erlbaum, 1993.

\bibitem{Anderson1989}
J.~R. Anderson and R.~Thompson.
\newblock {\em Use of Analogy in a Production System Architecture}, page
  267–297.
\newblock Cambridge University Press, USA, 1989.

\bibitem{chess1995}
M.~Bain and S.~H. Muggleton.
\newblock {\em Learning Optimal Chess Strategies}, pages 291--309.
\newblock Oxford University Press, Inc., New York, NY, USA, 1995.

\bibitem{barto1989learning}
A.~G. Barto, R.~S. Sutton, and C.~Watkins.
\newblock {\em Learning and sequential decision making}.
\newblock University of Massachusetts Amherst, MA, 1989.

\bibitem{bengio2009curriculum}
Y.~Bengio, J.~Louradour, R.~Collobert, and J.~Weston.
\newblock Curriculum learning.
\newblock In {\em Proceedings of the 26th Annual International Conference on
  Machine Learning}, pages 41--48, 2009.

\bibitem{berry1995implicit}
D.~C. Berry and D.~E. Broadbent.
\newblock {\em Implicit learning in the control of complex systems}, pages
  131--150.
\newblock Lawrence Erlbaum Associates, Inc., 1995.

\bibitem{BRATKO1995143}
I.~Bratko, T.~Urbančič, and C.~Sammut.
\newblock Behavioural cloning: Phenomena, results and problems.
\newblock {\em IFAC Proceedings Volumes}, 28(21):143--149, 1995.

\bibitem{Bruner1956}
J.~S. Bruner, J.~J. Goodnow, and G.~A. Austin.
\newblock {\em A study of thinking: With an appendix on language by Roger W.
  Brown.}
\newblock New York, NY: Wiley, 1956.

\bibitem{Carbonell1985}
J.~Carbonell.
\newblock Derivational analogy: A theory of reconstructive problem solving and
  expertise acquisition.
\newblock {\em Machine Learning}, 11:26, 1985.

\bibitem{Carpenter1990}
P.~Carpenter, M.~Just, and P.~Shell.
\newblock What one intelligence test measures: A theoretical account of the
  processing in the raven progressive matrices test.
\newblock {\em Psychological review}, 97:404--431, 1990.

\bibitem{chi2005complex}
M.~Chi and S.~Ohlsson.
\newblock {\em Complex Declarative Learning.}
\newblock Cambridge University Press, 2005.

\bibitem{Chomsky1965}
N.~Chomsky.
\newblock {\em Aspects of the theory of syntax}.
\newblock Cambridge: M.I.T. Press, 1965.

\bibitem{cropper2017}
A.~Cropper.
\newblock {\em Efficiently learning efficient programs}.
\newblock PhD thesis, Imperial College London, 2017.

\bibitem{cropper2016}
A.~Cropper and S.~H. Muggleton.
\newblock Metagol system.
\newblock https://github.com/metagol/metagol, 2016.

\bibitem{cropper2019}
A.~Cropper and S.~H. Muggleton.
\newblock Learning efficient logic programs.
\newblock {\em Machine Learning}, 108:1063--1083, 2019.

\bibitem{dienes1999theory}
Z.~Dienes and J.~Perner.
\newblock A theory of implicit and explicit knowledge.
\newblock {\em Behavioral and brain sciences}, 22(5):735--808, 1999.

\bibitem{dvzeroski2001}
S.~D{\v{z}}eroski, L.~De~Raedt, and K.~Driessens.
\newblock Relational reinforcement learning.
\newblock {\em Machine Learning}, 43:7--52, 2001.

\bibitem{gentner1985}
D.~Gentner and R.~Landers.
\newblock Analogical reminding: A good match is hard to find.
\newblock {\em Proceedings of the International Conference on Systems, Man and
  Cybernetics}, 1985.

\bibitem{GOLDMAN199520}
S.~A. Goldman and M.~J. Kearns.
\newblock On the complexity of teaching.
\newblock {\em Journal of Computer and System Sciences}, 50:20--31, 1995.

\bibitem{Hauser2002}
M.~D. Hauser, N.~Chomsky, and W.~T. Fitch.
\newblock The faculty of language: what is it, who has it, and how did it
  evolve?
\newblock {\em Science}, 298:1569--1579, 2002.

\bibitem{Hind2019}
M.~Hind, D.~Wei, M.~Campbell, N.~Codella, A.~Dhurandhar, and A.~e.~a.
  Mojsilovic.
\newblock Ted: Teaching ai to explain its decisions.
\newblock {\em Proceedings of the 2019 AAAI/ACM Conference on AI, Ethics, and
  Society}, 2019.

\bibitem{Hobbs2008}
J.~R. Hobbs.
\newblock {\em Abduction in Natural Language Understanding}.
\newblock Stevenage: Peter Peregrinus, 2008.

\bibitem{holyoak1987}
K.~J. Holyoak and K.~Koh.
\newblock Surface and structural similarity in analogical transfer.
\newblock {\em Memory \& Cognition 15(4)}, pages 332--340, 1987.

\bibitem{iyer2018transparency}
R.~Iyer, Y.~Li, H.~Li, M.~Lewis, R.~Sundar, and K.~Sycara.
\newblock Transparency and explanation in deep reinforcement learning neural
  networks.
\newblock In {\em Proceedings of the 2018 AAAI/ACM Conference on AI, Ethics,
  and Society}, pages 144--150, 2018.

\bibitem{Johnson_laird1986}
P.~N. Johnson-Laird.
\newblock {\em Mental Models: Towards a Cognitive Science of Language,
  Inference, and Consciousness}.
\newblock Harvard University Press, USA, 1986.

\bibitem{kahneman2011thinking}
D.~Kahneman.
\newblock {\em Thinking, fast and slow}.
\newblock Macmillan, 2011.

\bibitem{Kolmogorov1963}
A.~N. Kolmogorov.
\newblock On tables of random numbers.
\newblock {\em Sankhya: The Indian Journal of Statistics, Series A,},
  207(25):369–375, 1963.

\bibitem{Lemke1967}
E.~Lemke, H.~Klausmeier, and C.~Harris.
\newblock Relationship of selected cognitive abilities to concept attainment
  and information processing.
\newblock {\em Journal of educational psychology}, 58:27--35, 1967.

\bibitem{lin2014}
D.~Lin, E.~Dechter, K.~Ellis, J.~Tenenbaum, and S.~H. Muggleton.
\newblock Bias reformulation for one-shot function induction.
\newblock {\em In Proceedings of the 23rd European Conference on Artificial
  Intelligence (ECAI 2014)}, pages 525--530, 2014.

\bibitem{michie1963}
D.~Michie.
\newblock Experiments on the mechanization of game-learning part i.
  characterization of the model and its parameters.
\newblock {\em The Computer Journal, Volume 6, Issue 3}, pages 232--236, 1963.

\bibitem{michie1983}
D.~Michie.
\newblock Inductive rule generation in the context of the fifth generation.
\newblock {\em Machine Learning Workshop}, page~65, 1983.

\bibitem{Michie_cognitive_models}
D.~Michie, M.~Bain, and J.~Hayes-Michie.
\newblock {\em Cognitive models from sub cognitive skills.}, pages 71--99.
\newblock Peter Peregrinus, 1990.

\bibitem{Michie1992BuildingSR}
D.~Michie and R.~Camacho.
\newblock Building symbolic representations of intuitive real-time skills from
  performance data.
\newblock In {\em Machine Intelligence}, 1992.

\bibitem{Miller1956}
G.~A. Miller.
\newblock The magical number seven, plus or minus two: Some limits on our
  capacity for processing information.
\newblock {\em The Psychological Review}, 63:81--97, 1956.

\bibitem{miller2017explanation}
T.~Miller.
\newblock Explanation in artificial intelligence: Insights from the social
  sciences.
\newblock {\em Artificial Intelligence}, 267:1--38, 2019.

\bibitem{MILLER2017}
T.~Miller, P.~Howe, and L.~Sonenberg.
\newblock Explainable ai: Beware of inmates running the asylum or: How i learnt
  to stop worrying and love the social and behavioural sciences.
\newblock {\em Proc. IJCAI Workshop Explainable Artif. Intell. Melbourne,
  Australia.}, 2017.

\bibitem{Mitchell1982}
T.~M. Mitchell.
\newblock Generalization as search.
\newblock {\em Artificial Intelligence}, 18:203--226, 1982.

\bibitem{mnih2015}
V.~Mnih, K.~Kavukcuoglu, and D.~e.~a. Silver.
\newblock Human-level control through deep reinforcement learning.
\newblock {\em Nature}, 518:529--533, 2015.

\bibitem{US2018}
S.~Muggleton, U.~Schmid, C.~Zeller, A.~Tamaddoni-Nezhad, and T.~Besold.
\newblock Ultra-strong machine learning: comprehensibility of programs learned
  with ilp.
\newblock {\em Machine Learning}, 2018.

\bibitem{ILP1991}
S.~H. Muggleton.
\newblock Inductive logic programming.
\newblock {\em New Gen. Comput.}, 8:295–318, 1991.

\bibitem{muggleton2019}
S.~H. Muggleton and C.~Hocquette.
\newblock Machine discovery of comprehensible strategies for simple games using
  meta-interpretive learning.
\newblock {\em New Generation Computing}, 37:203--217, 2019.

\bibitem{muggleton2013}
S.~H. Muggleton and D.~Lin.
\newblock Meta-interpretive learning of higher-order dyadic datalog: Predicate
  invention revisited.
\newblock {\em In Proceedings of the 23rd International Joint Conference
  Artificial Intelligence}, pages 1551--1557, 2013.

\bibitem{muggleton2014}
S.~H. Muggleton, D.~Lin, N.~Pahlavi, and A.~Tamaddoni-Nezhad.
\newblock Meta-interpretive learning: application to grammatical inference.
\newblock {\em Machine Learning}, pages 25--49, 2014.

\bibitem{Newell1990}
A.~Newell.
\newblock {\em Unified Theories of Cognition}.
\newblock Harvard University Press, USA, 1990.

\bibitem{newell1981mechanisms}
A.~Newell and P.~S. Rosenbloom.
\newblock Mechanisms of skill acquisition and the law of practice.
\newblock {\em Cognitive skills and their acquisition}, 1(1981):1--55, 1981.

\bibitem{Novick1991}
L.~Novick and K.~Holyoak.
\newblock Mathematical problem solving by analogy.
\newblock {\em Journal of experimental psychology. Learning, memory, and
  cognition}, 17:398--415, 1991.

\bibitem{quinlan1983}
J.~Quinlan.
\newblock {\em Learning Efficient Classification Procedures and Their
  Application to Chess End Games}, pages 463--482.
\newblock Springer Berlin Heidelberg, Berlin, Heidelberg, 1983.

\bibitem{quinlan1987}
J.~R. Quinlan.
\newblock Simplifying decision trees.
\newblock {\em International Journal of Man-Machine Studies}, 27:221--234,
  1987.

\bibitem{Rafferty2016}
A.~N. Rafferty, E.~Brunskill, T.~L. Griffiths, and P.~Shafto.
\newblock Faster teaching via pomdp planning.
\newblock {\em Cognitive science}, pages 1290--1332, 2016.

\bibitem{reed1990}
S.~K. Reed, C.~C. Ackinclose, and A.~A. Voss.
\newblock Selecting analogous problems: Similarity versus inclusiveness.
\newblock {\em Memory \& Cognition 18(1)}, pages 83--98, 1990.

\bibitem{reed1991use}
S.~K. Reed and C.~A. Bolstad.
\newblock Use of examples and procedures in problem solving.
\newblock {\em Journal of Experimental Psychology: Learning, Memory, and
  Cognition}, 17(4):753, 1991.

\bibitem{Schmid2000}
U.~Schmid and J.~Carbonell.
\newblock Empirical evidence for derivational analogy.
\newblock {\em Proceedings of the 21st annual conference of the cognitive
  science society}, 2000.

\bibitem{schmid2020mutual}
U.~Schmid and B.~Finzel.
\newblock Mutual explanations for cooperative decision making in medicine.
\newblock {\em KI-K{\"u}nstliche Intelligenz}, 34(2):227--233, 2020.

\bibitem{Schmid11}
U.~Schmid and E.~Kitzelmann.
\newblock Inductive rule learning on the knowledge level.
\newblock {\em Cognitive Systems Research}, 12:237--248, 2011.

\bibitem{seger1994implicit}
C.~A. Seger.
\newblock Implicit learning.
\newblock {\em Psychological bulletin}, 115(2):163, 1994.

\bibitem{sequeira2020interestingness}
P.~Sequeira and M.~Gervasio.
\newblock Interestingness elements for explainable reinforcement learning:
  Understanding agents' capabilities and limitations.
\newblock {\em Artificial Intelligence}, 288:103367, 2020.

\bibitem{shapiro1982}
A.~Shapiro and T.~Niblett.
\newblock Automatic induction of classification rules for a chess endgame.
\newblock In M.~Clarke, editor, {\em Advances in Computer Chess}, volume~3,
  pages 73--91. Pergammon, Oxford, 1982.

\bibitem{shapiro1982algorithmic}
E.~Y. Shapiro.
\newblock Algorithmic program debugging. acm distinguished dissertation, 1982.

\bibitem{Shohamy1996}
E.~Shohamy.
\newblock {\em Performance and competence in second language acquisition},
  pages 138--151.
\newblock United Kingdom: Cambridge University Press, 1996.

\bibitem{Silver2016}
D.~Silver, A.~Huang, C.~J. Maddison, A.~Guez, L.~Sifre, and G.~e.~a. van~den
  Driessche.
\newblock Mastering the game of go with deep neural networks and tree search.
\newblock {\em Nature}, 529(7587):484--489, 2016.

\bibitem{simon1976}
H.~A. Simon and J.~R. Hayes.
\newblock The understanding process: Problem isomorphs.
\newblock {\em Cognitive Psychology 8}, pages 165--190, 1976.

\bibitem{stumpf2016}
S.~Stumpf, A.~Bussone, and D.~O’sullivan.
\newblock Explanations considered harmful? user interactions with machine
  learning systems.
\newblock In {\em Proceedings of the ACM SIGCHI Conference on Human Factors in
  Computing Systems (CHI)}, 2016.

\bibitem{telle2019teaching}
J.~A. Telle, J.~Hern{\'a}ndez-Orallo, and C.~Ferri.
\newblock The teaching size: computable teachers and learners for universal
  languages.
\newblock {\em Machine Learning}, 108:1653--1675, 2019.

\bibitem{Urbancic1994b}
T.~Urbančič and I.~Bratko.
\newblock Reconstructing human skill with machine learning.
\newblock {\em Proceedings of the 11th European Conference on Artificial
  Intelligence}, pages 498--502, 1994.

\bibitem{wang2019verbal}
X.~Wang, S.~Yuan, H.~Zhang, M.~Lewis, and K.~Sycara.
\newblock Verbal explanations for deep reinforcement learning neural networks
  with attention on extracted features.
\newblock In {\em 2019 28th IEEE International Conference on Robot and Human
  Interactive Communication (RO-MAN)}, pages 1--7. IEEE, 2019.

\bibitem{watkins1989}
C.~{Watkins}.
\newblock {\em Learning from Delayed Rewards}.
\newblock PhD thesis, 1989.

\bibitem{zahavy2016}
T.~Zahavy, N.~B. Zrihem, and S.~Mannor.
\newblock Graying the black box: Understanding dqns.
\newblock {\em Proceedings of the 33rd International Conference on Machine
  Learning}, 2016.

\bibitem{Zambaldi2019DeepRL}
V.~F. Zambaldi, D.~C. Raposo, A.~Santoro, V.~Bapst, Y.~Li, and I.~e.~a.
  Babuschkin.
\newblock Deep reinforcement learning with relational inductive biases.
\newblock In {\em ICLR}, 2019.

\bibitem{ZellerSchmid16}
C.~Zeller and U.~Schmid.
\newblock Automatic generation of analogous problems to help resolving
  misconceptions in an intelligent tutor system for written subtraction.
\newblock In {\em Workshops Proceedings for the Twenty-fourth International
  Conference on Case-Based Reasoning}, volume 1815, pages 108--117, 2016.

\bibitem{Zeller2017AHL}
C.~Zeller and U.~Schmid.
\newblock A human like incremental decision tree algorithm: Combining rule
  learning, pattern induction, and storing examples.
\newblock In {\em LWDA}, 2017.

\bibitem{Zhu2015}
X.~Zhu.
\newblock Machine teaching: An inverse problem to machine learning and an
  approach toward optimal education.
\newblock In {\em Proceedings of the Twenty-Ninth AAAI Conference on Artificial
  Intelligence}, page 4083–4087. AAAI Press, 2015.

\end{thebibliography}
